\documentclass[runningheads]{llncs}

\usepackage{eccv}

\usepackage{eccvabbrv}

\usepackage{graphicx}
\usepackage{booktabs}

\usepackage[accsupp]{axessibility}  

\usepackage{hyperref}

\usepackage{orcidlink}

\DeclareMathOperator*{\argmax}{argmax}
\DeclareMathOperator*{\argmin}{argmin}
\usepackage[table]{xcolor}
\usepackage{multirow}
\usepackage{multicol}
\usepackage{newunicodechar}
\newunicodechar{♡}{\ensuremath{\heartsuit}}
\usepackage{pgfplots}
\usepackage{color, colortbl}
\definecolor{cBlue}{HTML}{5383EC}
\definecolor{cRed}{HTML}{D85140}
\definecolor{cYellow}{HTML}{F1BF42}
\definecolor{cGreen}{HTML}{58A65C}
\definecolor{cOrange}{HTML}{ED762F}
\definecolor{cMint}{HTML}{69BBC4}
\definecolor{cCoOp}{HTML}{f2aa84}

\usepackage{amsmath}
\usepackage{amssymb}
\usepackage{algorithm}
\usepackage{algpseudocode}
\newcommand{\acc}[2]{\ensuremath{#1{\scriptstyle \pm #2}}}

\begin{document}

\title{AdaBoosting Text Prompts \\ for Vision-Language Models} 

\titlerunning{TPB}

\author{Seokhee Jin\inst{1}\textsuperscript{$\star$}\orcidlink{0009-0004-6845-4823} \and
Changhwan Sung\inst{2}\textsuperscript{$\star$}\orcidlink{0009-0009-7661-0925} \and
Sunung Mun\inst{2}\orcidlink{0009-0004-7507-2719} \and
Hoyoung Kim\inst{3}\orcidlink{0009-0001-3356-8626} \and
Jungseul Ok\inst{2}\textsuperscript{\dag}\orcidlink{0000-0003-4742-2473}}

\authorrunning{S.~Jin et al.}


\institute{
KT Corporation, Republic of Korea \and
Pohang University of Science and Technology (POSTECH), Republic of Korea \and
National AI Research Lab, Republic of Korea \\
\email{seokhee.jin@kt.com, \{changhwan.sung, mtablo, hoyoung.kim, jungseul.ok\}@postech.ac.kr} \\
\url{https://sung0503.github.io/TPB}
}

\maketitle

\begingroup
\renewcommand{\thefootnote}{}
\footnotetext{\textsuperscript{$\star$}These authors contributed equally to this work.} \footnotetext{\textsuperscript{\dag}Corresponding author.}
\endgroup

\begin{abstract}
The classification accuracy of pretrained Vision-Language Models (VLMs) relies on the quality of the text prompts.
Handcrafted templates and Large Language Model (LLM)-generated descriptions not only make predictions more interpretable, but also enable reuse of the same prompts across heterogeneous VLMs.
Recent works construct task-adapted text prompts with a small number of labeled images.
However, existing few-shot text prompting methods do not explicitly focus on misclassified examples during prompt construction, leading to only marginal improvements even as more shots become available.
To fully exploit few-shot supervision, we propose Text Prompt Boosting (TPB), an AdaBoost-inspired framework that treats each text-prompt-based classifier as a weak learner and sequentially aggregates them into a strong ensemble by explicitly targeting hard, misclassified examples.
Extensive experiments show that TPB preserves task-intrinsic, model-agnostic cues in text space, enabling robust cross-model transfer.
Across eleven classification benchmarks, TPB improves accuracy on the source model and preserves shot-driven gains when transferred to larger, more capable VLMs, where existing methods struggle to sustain such improvements.

\end{abstract}

\section{Introduction}
\label{sec:intro}
{
    \pgfplotsset{compat=1.18}
    \pgfplotsset{
        shotaxis/.style={
            height=3cm,
            width=\linewidth,
            label style={font=\tiny},
            tick label style={font=\tiny},
            xlabel=Shot,
            ylabel=Avg. Acc. (\%),
            xmin=-0.2, xmax=17.2,
            ymin=60, ymax=88,
            xtick={1, 2, 4, 8, 16},
            ytick={64, 68, 72, 76, 80, 84},
            xlabel style={yshift=0.10cm},
            ylabel style={yshift=-0.15cm},
            legend style={
                at={(1,0)},
                anchor=south east,
                draw=none,
                fill=none,
                font=\tiny,
                legend cell align=left,
                row sep=-2pt,
                scale=0.1,
                transform shape,
                /tikz/every node/.append style={scale=0.8, transform shape},
            },
            legend columns=1,
        }
    }

    \begin{figure}[t]
        \centering
        
        \hfill
        \begin{subfigure}[b]{0.6\linewidth}
            \centering
            \includegraphics[height=4cm, keepaspectratio]{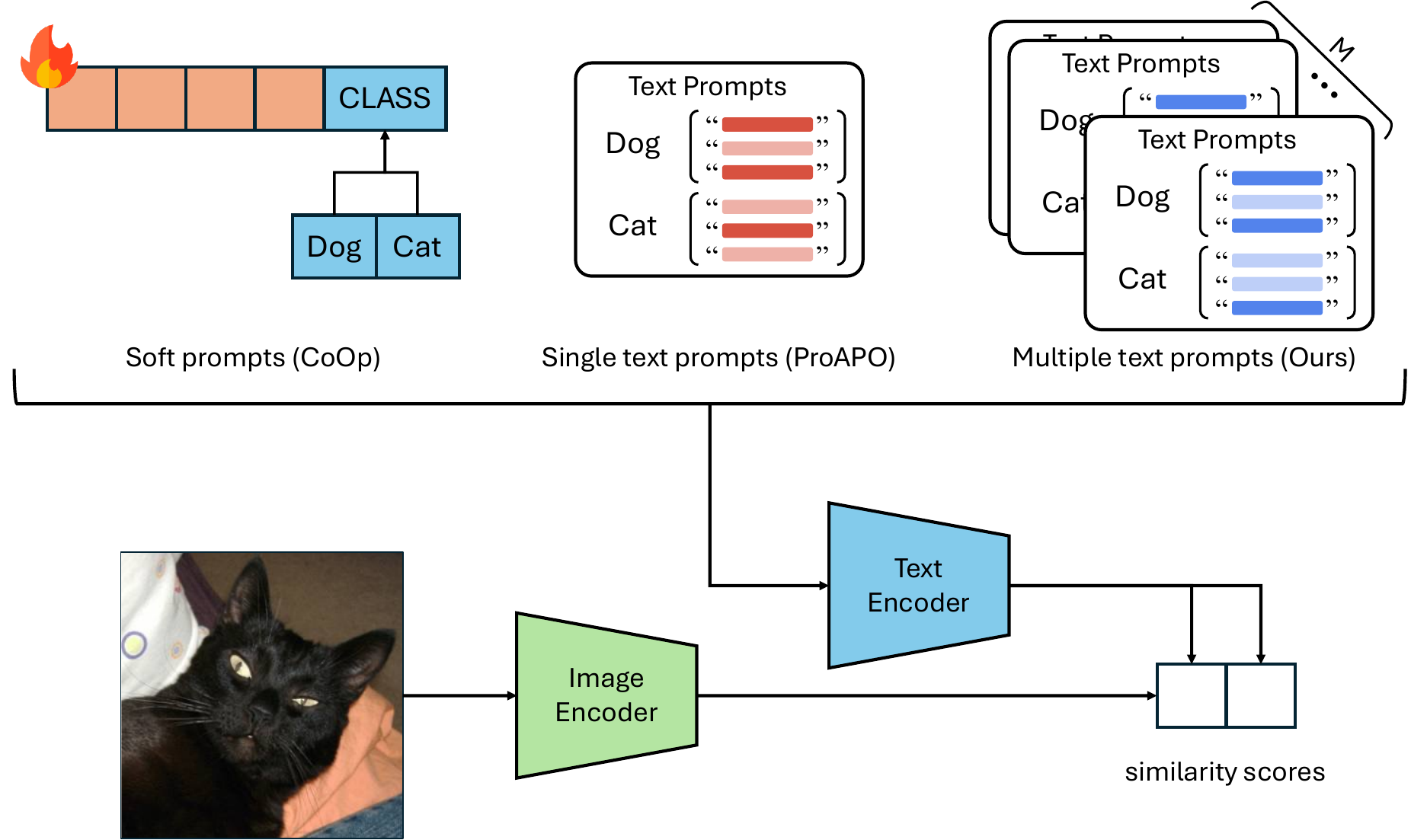}
            \caption{VLM-based inference}
            \label{fig:fig1a}
        \end{subfigure}
        \hfill
        \hfill
        \hfill
        \begin{subfigure}[b]{0.3\linewidth}
            \centering
            \begin{tikzpicture}[trim axis left, trim axis right]
                \begin{axis}[
                    shotaxis,
                    scale only axis,
                ]
                    \addplot[
                        cBlue, mark=*, thick, densely dotted, mark options={solid},
                        legend image code/.code={
                            \draw[cBlue, thick, solid] (0cm,0.03cm) -- (0.34cm,0.03cm);
                            \draw[cBlue, thick, densely dotted] (0cm,-0.01cm) -- (0.34cm,-0.01cm);
                            \draw[cBlue, thick, solid]
                                plot[only marks, mark=*, mark size=1.4pt, mark options={solid}]
                                coordinates {(0.17cm,0.005cm)};
                        }
                    ]
                        coordinates {(0,0)};
                    \addplot[
                        cRed, thick, densely dotted, mark options={solid},
                        legend image code/.code={
                            \draw[cRed, thick, solid] (0cm,0.03cm) -- (0.34cm,0.03cm);
                            \draw[cRed, thick, densely dotted] (0cm,-0.01cm) -- (0.34cm,-0.01cm);
                            \draw[cRed, thick, solid]
                                plot[only marks, mark=triangle*, mark size=1.4pt, mark options={solid}]
                                coordinates {(0.17cm,0.005cm)};
                        }
                    ]
                        coordinates {(0,0)};
                    \addplot[
                        cOrange, thick, densely dotted, mark options={solid},
                        legend image code/.code={
                            \draw[cOrange, thick, densely dotted] (0cm,0cm) -- (0.34cm,0cm);
                            \draw[cOrange, thick, solid]
                                plot[only marks, mark=square*, mark size=1.2pt, mark options={solid}]
                                coordinates {(0.17cm,0.005cm)};
                        }
                    ]
                        coordinates {(0,0)};
                    \addplot[cOrange, mark=square*, mark size=0.8pt, very thick, densely dotted, mark options={solid}]
                        coordinates {(1, 62.88) (2, 65.34) (4, 69.69) (8, 72.88) (16, 75.81)};
                    \addplot[cRed, mark=triangle*, mark size=1pt, very thick, densely dotted, mark options={solid}]
                        coordinates {(1, 66.07) (2, 66.55) (4, 66.97) (8, 67.17) (16, 67.35)};
                    \addplot[cBlue, mark=*, mark size=0.95pt, very thick, densely dotted, mark options={solid}]
                        coordinates {(1, 67.07) (2, 68.53) (4, 70.29) (8, 71.85) (16, 72.87)};
                    \addplot[cRed, mark=triangle*, mark size=1pt, very thick, mark options={solid}]
                        coordinates {(1, 79.59) (2, 79.53) (4, 79.87) (8, 79.93) (16, 80.01)};
                    \addplot[cBlue, mark=*, mark size=0.95pt, very thick, mark options={solid}]
                        coordinates {(1, 79.66) (2, 80.35) (4, 81.05) (8, 81.88) (16, 82.07)};
             
                    \node[anchor=north west, font=\tiny, text=black] at (rel axis cs:0.01, 0.98) {ViT-B/32 $\rightarrow$ ViT-L};
                    \legend{\textbf{TPB (Ours)}, ProAPO, CoOp}
                \end{axis}
            \end{tikzpicture}
            \caption{Shot scalability}
            \label{fig:combined_results}
            \label{fig:fig1b}
        \end{subfigure}
        \hfill
        \hfill

        \caption{
            \textbf{Concept of TPB and its transfer-robust shot scalability.}
            (a) While soft prompting (\eg, CoOp) optimizes continuous prompt vectors via backpropagation and text-based baselines (\eg, ProAPO) rely on a single prompt set, TPB iteratively builds a collection of prompt banks to form a strong classifier.
            (b) Performance comparison upon cross-model transfer from ViT-B/32 to five ViT-L-scale VLMs, where dashed and solid lines indicate source and transfer performance, respectively.
            While CoOp cannot be directly transferred and ProAPO fails to preserve shot-driven gains, TPB maintains these shot-driven gains while demonstrating superior scalability on the source model.
        }

            
        \label{fig:motivation}
    \end{figure}
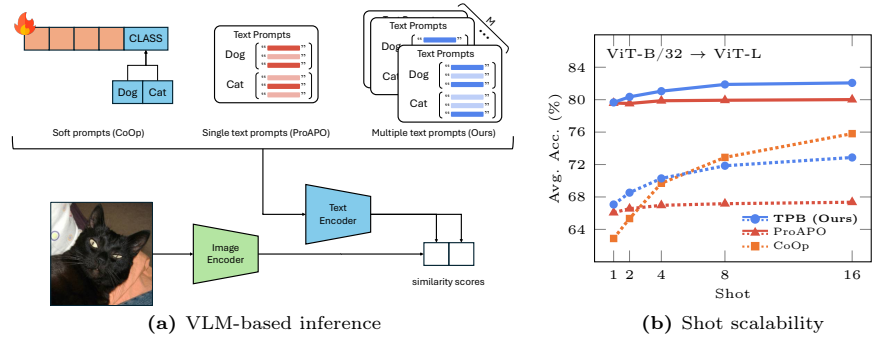
}

Vision-Language Models (VLMs)~\cite{radford2021learning, jia2021align, cherti2023openclip, li2023clipa, tschannen2025siglip2, chuang2025metaclip2, fang2023dfn, sun2023evaclip} such as CLIP~\cite{radford2021learning} enable zero-shot image classification using natural-language text prompts that describe each class.
In practice, the classification accuracy is sensitive to the quality of the prompts.
To avoid labor-intensive manual prompt engineering, prior studies~\cite{menon2022dclip, pratt2023cupl, roth2023waffling, wu2023gpt4vis, ren2023chatgpt} have explored LLM-based automated prompt generation.
A common practice is to query the Large Language Model with a generic instruction (\eg, ``\emph{What does a \{class\} look like?}'')~\cite{pratt2023cupl}, resulting in prompts based solely on the LLM’s prior knowledge.
However, such prompts often lack proper grounding in the target visual distribution and suffer from inaccuracies due to class-name ambiguity or hallucinations.

Recent studies~\cite{liu2024llmbo, qu2025proapo} address this lack of visual grounding by leveraging a small number of labeled images to facilitate the discovery of task-optimized prompts.
To exploit these few-shot examples, these methods typically optimize a single aggregate metric (\eg, overall validation accuracy) as their primary objective for prompt generation.
However, the fundamental limitation is that such a metric can be dominated by a subset of easily recognizable samples.
Once a candidate prompt correctly classifies these trivial instances, validation accuracy saturates prematurely, failing to capture the broader visual diversity of the target domain.
As a result, these methods often overfit to the few-shot set and yield only marginal gains even as more labeled shots become available.

To address these limitations and effectively exploit few-shot supervision, we propose Text Prompt Boosting (TPB), drawing inspiration from the classic AdaBoost algorithm.
Rather than optimizing a single prompt that quickly plateaus on easy samples, TPB iteratively builds an ensemble of natural-language prompts (as shown in \Cref{fig:fig1a}) by explicitly redirecting focus onto ``hard examples.''
At each boosting round, TPB reweights misclassified training examples, effectively forcing the selection of new prompts from an LLM-generated pool that resolve these specific errors.
By systematically addressing failure modes, TPB maximizes the utility of few-shot examples to cover broader visual variations, successfully capturing model-agnostic task knowledge.

Extensive evaluations across diverse image classification benchmarks show that TPB consistently outperforms state-of-the-art baselines.
\Cref{fig:fig1b} previews this transfer-robust shot scalability across five ViT-L-scale VLMs.
By effectively exploiting few-shot supervision, TPB exhibits superior shot scalability on the source model.
Furthermore, although natural-language prompts are inherently transferable across heterogeneous VLMs through simple re-embedding, prior methods often fail to retain shot-driven gains after transfer.
In contrast, TPB preserves the shot-driven improvements observed on the source model when transferred to larger, more capable VLMs, maintaining clear performance gains even at higher shot counts.

Our main contributions are summarized as follows:
\begin{itemize}
    \item[\textbullet] We propose a few-shot prompting framework that combines AdaBoost with natural-language text prompts, yielding a boosted ensemble with significantly improved shot scalability over prior text-based methods.
    \item[\textbullet] We show that the prompt ensembles transfer seamlessly across heterogeneous VLMs by simple re-embedding, achieving superior cross-model performance compared to existing transfer baselines.
    \item[\textbullet] We conduct extensive experiments across standard few-shot benchmarks and multiple VLM backbones, accompanied by detailed component analyses.
\end{itemize}

\section{Related Work}
\label{sec:related_work}

\subsection{Prompt Adaptation of VLMs.}
\label{subsec:prompt_adaptation}
Prompting strategies for vision-language models broadly fall into soft (continuous) and hard (discrete) paradigms.
While soft prompting, often referred to as prompt-learning, which attaches learnable continuous context vectors to class names for parameter-efficient adaptation~\cite{zhou2022coop,zhou2022cocoop,khattak2023maple,khattak2023promptsrc,park2024prompt}, the learned prompts lie in a model-specific representation space, making them unreadable and difficult to transfer directly across models.
In contrast, hard prompt optimization searches over discrete natural-language tokens, ensuring that the resulting prompts remain human-readable and reusable across heterogeneous CLIP-style encoders.
To efficiently navigate the enormous language space, recent methods leverage LLMs to generate and refine prompts. For instance, DCLIP~\cite{menon2022dclip} and CuPL~\cite{pratt2023cupl} construct task-specific descriptions, whereas further methods incorporate few-shot adaptation via conversational feedback in LLMbo~\cite{liu2024llmbo} and evolution-based search in ProAPO~\cite{qu2025proapo}.
Alternatively, PEZ~\cite{wen2023hard} employs a gradient-based hard-prompt optimization technique.
However, despite the advantages of hard prompting, most existing methods lack mechanisms to effectively leverage more than a single labeled example per class.
As a result, their performance tends to saturate and improves only marginally, even when more labeled examples are provided.
Our work follows the hard-prompt paradigm but introduces an AdaBoost-based framework to effectively exploit additional supervision while preserving interpretability and cross-model transfer.

\subsection{Diverse Applications of AdaBoost.}
\label{subsec:adaboost_application}
Adaptive Boosting (AdaBoost)~\cite{freund1997adaboost,zhu2005multi,friedman2000additive} sequentially combines many weak classifiers by iteratively reweighting training examples, emphasizing those misclassified by earlier classifiers. This general framework has been applied across diverse domains. In the computer vision domain, AdaBoost has been widely used in fast detection pipelines, such as cascaded detectors~\cite{viola2001rapid} and integral channel feature-based detectors~\cite{dollar2009integral}. In text classification and information retrieval, boosting aggregates weak lexical predicates into confidence-rated document classifiers \cite{schapire2000boostexter} and optimizes pairwise orderings for ranking \cite{freund2003efficient}. Unlike these applications, we address cross-modal image recognition with VLMs, treating class-wise text prompts as weak classifiers that map an image to class-wise similarity scores.

\section{Method}
\label{sec:method}
\begin{figure}[t]
    \centering
    \includegraphics[width=\linewidth]{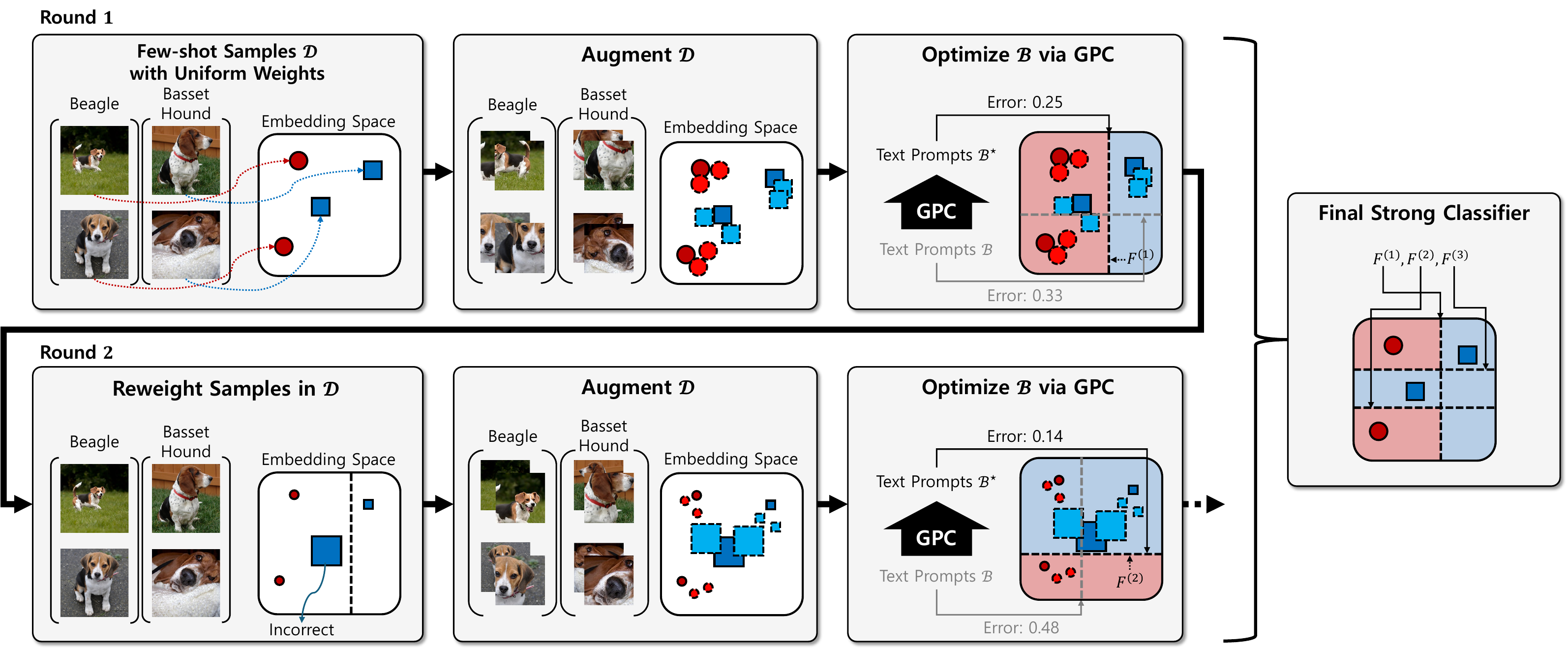}
    \caption{
        \textbf{Overview of our proposed TPB framework.}
        Initially, sample weights are uniformly distributed on the few-shot set $\mathcal{D}$.
        At each round $m$, weights are updated based on the previous round's weak classifier.
        Then, we apply augmentation and an optimal prompt collection $\mathcal{B}^\star$ is derived via GPC; $\mathcal{B}^\star$ defines the current weak classifier.
        After the final round, all weak classifiers are aggregated into a single strong classifier.
    }
    \label{fig:overview}
\end{figure}

We propose Text Prompt Boosting (TPB), a prompt-ensemble framework for shot-scalable and transfer-robust few-shot adaptation of pretrained VLMs.
TPB treats a class-wise prompt collection as a text-based weak classifier and builds a strong classifier via an AdaBoost-style ensemble over multiple boosting rounds.
We begin with preliminaries in \Cref{subsec:prelim}, then introduce large prompt pools in \Cref{subsec:largepools} and our weak learner, Greedy Prompt Composition (GPC), in \Cref{subsec:gpc}.
Finally, we describe how GPC is integrated into the TPB boosting loop in \Cref{subsec:tpb}.
An overview of TPB is shown in \Cref{fig:overview}.

\subsection{Preliminaries}
\label{subsec:prelim}
\subsubsection{VLM Classification with Text Prompts.}
\label{subsubsec:clip_inference}
We consider $K$-way classification for an image $\mathbf{x}$.
Given a pretrained VLM with image encoder $E_{\mathrm{img}}$ and text encoder $E_{\mathrm{text}}$, we form a CLIP-style linear classifier whose weights are text features derived from class prompts $t_k$.
Specifically, the logit for class $k$ is computed by cosine similarity between $E_{\mathrm{img}}(\mathbf{x})$ and $E_{\mathrm{text}}(t_k)$:
\begin{equation}
    s(\mathbf{x}, t_k) := \cos \bigl( E_{\mathrm{img}}(\mathbf{x}) , E_{\mathrm{text}}(t_k) \bigr) \, .
    \label{eq:single_sim}
\end{equation}
Instead of relying on a single prompt, multiple prompts can be grouped into a class-wise prompt bank $B_k = \{ t_{k,1}, \dots, t_{k,n_k} \}$, and compute the class score by simple averaging:
\begin{equation}
    s(\mathbf{x}, B_k) := \frac{1}{\lvert B_k \rvert} \sum_{t_k \in B_k} s(\mathbf{x}, t_k) \, .
    \label{eq:avg_sim}
\end{equation}
Let $\mathcal{B} := \{B_1, \ldots, B_K\}$ denote the collection of class-wise prompt banks.
Inference on an image $\mathbf{x}$ is then performed using this prompt collection with a fixed temperature $\tau > 0$:
\begin{equation}
    p(y = k \mid \mathbf{x}; \mathcal{B}) :=
    \frac{\exp \bigl(s(\mathbf{x}, B_k)/\tau\bigr)}
         {\sum_{i=1}^{K}\exp\bigl(s(\mathbf{x}, B_i)/\tau\bigr)} \, .
    \label{eq:prob_scalar}
\end{equation}
Based on these probabilities, the predicted class label is determined by selecting the class with the highest probability:
\begin{equation}
    \hat{y}(\mathbf{x}; \mathcal{B}) := \argmax_{k\in\{1,\dots,K\}} p(y = k \mid \mathbf{x}; \mathcal{B}) \, .
    \label{eq:vlm_pred}
\end{equation}

While prior text-prompting studies~\cite{menon2022dclip,pratt2023cupl,roth2023waffling,liu2024llmbo,qu2025proapo} construct only a single collection $\mathcal{B}$, our TPB framework sequentially constructs multiple diverse collections $\{\mathcal{B}^{(m)}\}_{m=1}^{M}$.
Each collection defines a weak classifier in an AdaBoost-style ensemble, enabling us to explicitly target hard examples while preserving the model transferability of discrete text prompts.

\subsubsection{AdaBoost Overview.}
\label{subsubsec:adaboost_overview}
AdaBoost~\cite{freund1997adaboost} is a prominent ensemble algorithm that combines weak learners into a strong learner by iteratively training weak classifiers on reweighted data.
Unlike standard AdaBoost, which typically uses decision trees as weak learners, our framework uses the VLM classifier induced by a prompt collection $\mathcal{B}$ as the weak learner.
At each boosting round $m \in \{1,\ldots,M\}$, we construct a prompt collection $\mathcal{B}^{(m)}$.
Since the weak classifier is fully specified by the VLM predictions induced by $\mathcal{B}^{(m)}$, the objective at round $m$ is to minimize the weighted classification error on the training set $\mathcal{D}=\{(\mathbf{x}_i,y_i)\}_{i=1}^{n}$:
\begin{equation}
    \mathcal{L}^{(m)} := \sum_{i=1}^{n} w_i^{(m)} \,\mathbb{I}\!\left[ 
        y_i \neq \hat{y}(\mathbf{x}_i; \mathcal{B}^{(m)}) 
    \right] \, ,
    \label{eq:weighted_error}
\end{equation}
where $w_i^{(m)}$ is the weight assigned to the $i$-th image at round $m$, indicating how much the current prompt collection should focus on it, and $\hat{y}$ denotes the predicted class defined in \Cref{eq:vlm_pred}.

After obtaining $\mathcal{B}^{(m)}$, we update the sample weights for the next round by increasing the weights of misclassified (hard) samples and decreasing those of correctly classified ones.
This fitting and reweighting process repeats, encouraging subsequent prompt collections to prioritize the remaining difficult examples.
The final prediction is obtained by aggregating the outputs from all $M$ rounds.
For example, aggregation can be performed via weighted majority voting:
\begin{equation}
    \hat{y}_{\mathrm{final}} = \operatorname*{arg\,max}_{k} \sum_{m=1}^{M} \alpha^{(m)} \cdot
    \mathbb{I}\!\left[ \hat{y}(\mathbf{x}; \mathcal{B}^{(m)}) = k \right] \, ,
    \label{eq:final_pred_hard_voting}
\end{equation}
where $\alpha^{(m)}$ is the weight for the $m$-th weak learner.
The specific reweighting and aggregation rules depend on the AdaBoost variant; we adopt SAMME.R~\cite{zhu2005multi}, a multi-class real-valued extension, which we refer to simply as AdaBoost in the rest of this paper.

\subsection{Large Prompt Pools}
\label{subsec:largepools}
Before initiating the boosting loop, we first construct a comprehensive prompt pool $\mathcal{P}=\{P_1,\dots,P_K\}$ for each class $k\in\{1,\ldots,K\}$.
To ensure sufficient expressive power, these prompt pools encompass both simple template-based prompts (\eg, ``\emph{a photo of a beagle.}'') and richer, more detailed sentences generated by LLMs (\eg, ``\emph{beagles have large, floppy ears that hang down to the sides of their face.}'').
Concretely, we utilize the 80 hand-crafted templates provided by CLIP~\cite{radford2021learning}, alongside LLM-generated sentences released by several prior studies~\cite{menon2022dclip,pratt2023cupl,wu2023gpt4vis,saha2024improved,zhu2024awt}.
To further expand the discrete hypothesis space and enrich semantic diversity, we also append concatenations of these templates and descriptive sentences to each pool.

\subsection{Greedy Prompt Composition}
\label{subsec:gpc}
At each boosting round $m\in\{1,\dots,M\}$, our framework constructs a collection $\mathcal{B}^{(m)} = \{B_k^{(m)}\}_{k=1}^K$ of class-specific prompt banks, where each $B_k^{(m)}$ is selected from the class-specific prompt pool $P_k$.
Exhaustively evaluating all joint prompt combinations over the $K$ classes is computationally infeasible; thus, we adopt a two-stage greedy strategy, which we refer to as \emph{Greedy Prompt Composition (GPC)}.
For notational simplicity, we omit the round index in this subsection and write $B_k$ and $w_i$ instead of $B_k^{(m)}$ and $w_i^{(m)}$.

\subsubsection{Stage 1: Single-Template Initialization.}
\label{subsubsec:stage1}
We initialize the prompt collection $\mathcal{B}=\{B_k\}_{k=1}^K$ with a single shared template.
Among the 80 standard templates $\Phi_{\mathrm{template}}=\{\phi_1,\dots,\phi_{80}\}$, we select $\phi^\star$ that minimizes the weighted classification error.
Concretely, for each candidate template $\phi\in\Phi_{\mathrm{template}}$, we form a template-induced collection $\mathcal{B}(\phi)=\{B_k(\phi)\}_{k=1}^K$ by setting $B_k(\phi)=[\,\phi(c_k)\,]$, where $c_k$ denotes the label text of class $k$.
We then compute the weighted error of the resulting classifier (under the current example weights) and choose the best template.
If multiple templates attain the same minimum error, we break ties randomly.
Accordingly, we define
\begin{equation}
    \mathcal{B}_{\mathrm{init}} := \{B_k\}_{k=1}^K,
    \quad
    B_k := [\, \phi^\star(c_k) \,] \ \ \forall k \in \{1,\dots,K\}.
\end{equation}
We adopt this single-template initialization for two reasons.
First, any classifier requires at least one prompt per class, and a shared template provides this with minimal overhead by simply inserting the class name.
Second, using one template consistently across classes already yields a reasonably strong baseline, which serves as a stable starting point for greedy refinement and promotes fast convergence.

\subsubsection{Stage 2: Class-wise Greedy Prompt Insertion.}
\label{subsubsec:stage2}
Starting from the single-template initialization $\mathcal{B}_{\mathrm{init}}$, we greedily expand the prompt banks by iteratively adding prompts to each class.
Specifically, given the current prompt bank collection $\mathcal{B}$ and class $k$, we evaluate each candidate prompt $t \in P_k$ by temporarily appending it to the bank, \ie,  $\widetilde{B}_k(t) = B_k \| [t]$, and computing the resulting weighted error.
Note that the class score is computed as an unweighted average over the prompts in a bank as defined in \Cref{eq:avg_sim}.
Under this formulation, adding a single prompt to an already populated bank may induce only a small change in the class score, which can be insufficient to resolve the remaining hard examples.
Rather than introducing learnable continuous weights, which would require gradient updates, we implicitly approximate continuous weighting within a strictly discrete space by permitting duplicate selections of the same prompt; since the score is an unweighted average, duplicating a prompt increases its effective contribution.
Ultimately, we select the candidate $t_k^\star$ that yields the largest reduction in the weighted error, updating $B_k \leftarrow B_k \| [t_k^\star]$ if such a candidate exists, or leaving $B_k$ unchanged otherwise.
We repeat this procedure cyclically across all classes $k = 1, \dots, K$ until a full pass over all classes yields no further updates, resulting in the optimized bank collection $\mathcal{B}^\star$.

\subsection{Integration into the TPB Framework}
\label{subsec:tpb}

\subsubsection{Boosting-Loop Image Augmentation.}
\label{subsec:augment}
To induce informative variations and prevent overfitting, which occurs when the ensemble quickly memorizes the limited samples, we apply image augmentation to the few-shot dataset $\mathcal{D}=\bigl\{(\mathbf{x}_i,y_i)\bigr\}_{i=1}^n$ before executing the GPC algorithm. With an augmentation factor $a \in \mathbb{N}$, we generate randomly transformed copies for each training image $\mathbf{x}_i$ using simple operations (\eg, random resized crop and horizontal flip), resulting in an augmented dataset $\mathcal{D}_{\mathrm{aug}}$ of size $an$. We replicate the original sample weight $w_i$ for each copy and renormalize the weight vector $\mathbf{w}_{\mathrm{aug}}$ to sum to one. Consequently, the weighted errors evaluated during the GPC stages are computed over this augmented dataset $\mathcal{D}_{\mathrm{aug}}$ and its corresponding weights $\mathbf{w}_{\mathrm{aug}}$.

\subsubsection{Iterative Weight Update and Final Ensemble.}
\label{subsubsec:final_tpb}
In the overall TPB framework, the prompt bank collection $\mathcal{B}^{\star(m)}$ obtained from GPC serves as a weak classifier for boosting round $m$.
We then apply this weak classifier to the original (non-augmented) few-shot dataset $\mathcal{D}$ to update the sample weights $\mathbf{w}^{(m)}$ to $\mathbf{w}^{(m+1)}$, increasing the weights of misclassified examples and decreasing those of correctly classified ones.
Using these updated weights, the next boosting round constructs a new weak classifier that focuses more intensely on the previously misclassified samples.
Note that we initialize the weights uniformly in the first round, i.e., $w_i^{(1)} = 1/|\mathcal{D}|$ for all $(\mathbf{x}_i,y_i)\in\mathcal{D}$.
After the final round $M$, all weak classifiers are aggregated into a strong classifier. For the reweighting and aggregation rules, we adopt SAMME.R~\cite{zhu2005multi}, a multi-class variant of AdaBoost.
Detailed pseudocodes are provided in the supplementary material.

\section{Experiments}
\label{sec:exp}

\subsection{Experimental Setup}
\subsubsection{Datasets.}
Following prior text prompting ~\cite{menon2022dclip, pratt2023cupl,liu2024llmbo, qu2025proapo} and prompt-learning studies~\cite{zhou2022coop,khattak2023promptsrc}, we consider eleven image classification datasets that cover a wide range of domains and granularity: ImageNet-1K~\cite{deng2009imagenet} and Caltech101~\cite{fei2004caltech} for generic object classification, OxfordPets~\cite{parkhi2012pets}, StanfordCars~\cite{krause2013cars}, Flowers102~\cite{nilsback2008flowers}, Food101~\cite{bossard2014food}, FGVCAircraft~\cite{maji2013aircraft} for fine-grained image classification, SUN397~\cite{xiao2010sun} for scene recognition, DTD~\cite{cimpoi2014dtd} for texture classification, EuroSAT~\cite{helber2019eurosat} for satellite image classification, and UCF101~\cite{soomro2012ucf101} for action classification.

\subsubsection{Implementation Details.}
The number of boosting rounds $M$ is set to $50$, except for ImageNet, where it is set to $30$, and the augmentation factor $a$ is set to $4$.
Our framework is implemented based on the AdaBoost implementation in scikit-learn~\cite{scikit-learn}. For the temperature parameter $\tau$ in \Cref{eq:prob_scalar}, we use a fixed value $\tau = 1$.
Note that during inference, the text embeddings of the ensembled prompts can be pre-computed.
Thus, for any new image, TPB only requires a single forward pass, ensuring that our method does not incur significant latency overhead compared to other baselines.
All results, except for LLMbo~\cite{liu2024llmbo} as described in~\Cref{subsubsec:baselines}, are averaged over three random seeds; full shot-wise and target-wise tables with standard deviations are in the supplementary material.

\subsection{Few-Shot Adaptation and Shot Scalability}
{
    \newcommand{\myangle}{90}
    \newcommand{\gr}{\color{gray!70}}
    \begin{table}[t]
        \caption{
            \textbf{Comparison of shot scalability among text-based methods.} Top-1 accuracy ($\%$) on eleven datasets using OpenAI CLIP RN50. TPB outperforms all text-based baselines in both one-shot and sixteen-shot scenarios.
        }
        \label{tab:mainshot}
        
        \centering
        \footnotesize
        \setlength{\tabcolsep}{2pt}

        \scalebox{0.9}{
        \begin{tabular}{cl|ccccccccccc|c}
            \toprule
            \textbf{Shot} & \textbf{Method} & \rotatebox{\myangle}{\textbf{IN-1K}} & \rotatebox{\myangle}{\textbf{Caltech}} & \rotatebox{\myangle}{\textbf{Pets}} & \rotatebox{\myangle}{\textbf{Cars}} & \rotatebox{\myangle}{\textbf{Flowers}} & \rotatebox{\myangle}{\textbf{Food}} & \rotatebox{\myangle}{\textbf{Aircraft}} & \rotatebox{\myangle}{\textbf{SUN}} & \rotatebox{\myangle}{\textbf{DTD}} & \rotatebox{\myangle}{\textbf{ESAT}} & \rotatebox{\myangle}{\textbf{UCF}} & \rotatebox{\myangle}{\textbf{Avg.}} \\
            
            \midrule
            \rowcolor{gray!15}
            \multicolumn{14}{c}{\textit{Zero-shot scenario}}\\
            \midrule
            
            \multirow{3}{*}{\rotatebox{90}{ZS}} & CLIP & 58.3 & 85.8 & 83.7 & 55.9 & 61.2 & 75.2 & 14.5 & 58.5 & 40.0 & 24.2 & 58.4 & 56.0 \\
            & DCLIP & 59.6 & 88.6 & 83.1 & 53.9 & 66.3 & 76.6 & 16.8 & 61.0 & 41.7 & 37.6 & 60.8 & 58.7 \\
            & CuPL & 61.5 & 88.0 & 87.6 & 56.1 & 68.1 & 77.1 & 18.3 & 61.9 & 47.6 & 36.3 & 62.1 & 60.4 \\
            
            \midrule
            \rowcolor{gray!15}
            \multicolumn{14}{c}{\textit{Few-shot scenario}}\\
            \midrule
            
            \multirow{5}{*}{\rotatebox{90}{1 shot}}
            & CoOp & 55.5 & 88.0 & 86.3 & 55.6 & 68.2 & 74.2 & 8.4 & 60.2 & 43.3 & 51.1 & 61.9 & 59.3 \\
            \cmidrule{2-14}
            & PEZ & 34.2 & 66.5 & 68.8 & 38.2 & 57.8 & 51.9 & 12.8 & 35.6 & 28.8 & 29.4 & 41.8 & 42.4 \\
            & LLMbo & 59.6 & \textbf{89.1} & \underline{88.1} & 56.2 & 67.2 & \textbf{78.3} & 18.1 & 61.0 & 44.8 & 49.0 & 60.2 & 61.1 \\
            & ProAPO & \textbf{61.1} & \underline{89.0} & \textbf{89.0} & \textbf{57.6} &   \underline{68.9} & \underline{78.3} & \underline{18.2} & \underline{61.9} & \underline{48.3} & \textbf{53.3} & \underline{63.6} & \underline{62.7} \\
            \rowcolor{blue!8}
            \cellcolor{white} & \textbf{TPB (Ours)} & \underline{60.7} & 88.8 & 87.2 & \underline{57.4} & \textbf{77.0} & 76.4 & \textbf{19.7} & \textbf{63.7} & \textbf{53.1} & \underline{51.5} & \textbf{66.2} & \textbf{63.8} \\
            
            \midrule
            
            \multirow{5}{*}{\rotatebox{90}{16 shot}}
            & CoOp & 60.2 & 91.9 & 86.3 & 72.9 & 94.7 & 74.4 & 31.5 & 68.5 & 63.4 & 83.0 & 75.7 & 73.0 \\
            \cmidrule{2-14}
            & PEZ & 54.5 & 85.6 & 82.3 & 56.9 & \underline{75.5} & 70.5 & \underline{20.6} & 57.0 & 50.6 & 58.1 & 60.4 & 61.1 \\
            & LLMbo & 59.9 & 89.5 & 88.3 & 56.8 & 67.4 & 78.3 & 18.1 & 60.8 & 44.9 & 51.4 & 60.5 & 61.4 \\
            & ProAPO & \underline{60.1} & \underline{90.1} & \textbf{89.1} & \underline{58.5} & 73.3 & \underline{78.8} & 18.7 & \underline{63.0} & \underline{52.5} & \underline{58.6} & \underline{65.5} & \underline{64.4} \\
            \rowcolor{blue!8}
            \cellcolor{white} & \textbf{TPB (Ours)} & \textbf{64.7} & \textbf{91.8} & \underline{89.0} & \textbf{65.0} & \textbf{85.9} & \textbf{79.0} & \textbf{24.8} & \textbf{70.7} & \textbf{62.8} & \textbf{66.9} & \textbf{73.0} & \textbf{70.3} \\
            \bottomrule
        \end{tabular}
        }
    \end{table}
}

\subsubsection{Text-based Baselines.}
\label{subsubsec:baselines}
For zero-shot baselines, we employ the standard template ``\emph{a photo of a \{class\}.}'' (denoted as CLIP), along with DCLIP~\cite{menon2022dclip} and CuPL~\cite{pratt2023cupl}.
For few-shot baselines, we compare our method against LLMbo~\cite{liu2024llmbo}, ProAPO~\cite{qu2025proapo}, and a gradient-based hard-prompt optimization method, PEZ~\cite{wen2023hard}.
Except for LLMbo, all baseline results are reproduced using the authors' official implementations.
Since PEZ was not originally designed for our task, we reimplement it for image classification; implementation details are provided in the supplementary material.

\subsubsection{Shot Scalability.}
\Cref{tab:mainshot} compares our TPB with existing text-based methods on eleven datasets using OpenAI CLIP~\cite{radford2021learning} with a ResNet-50 backbone~\cite{he2016deep}. While other text-based baselines show only marginal or no gains as more shots become available, our TPB achieves both strong few-shot performance and clear improvements with additional labeled data. At one shot, TPB slightly outperforms ProAPO by about 1.1 percentage points (pp) in average accuracy. When the number of shots increases from one to sixteen, TPB gains 6.5 pp and outperforms ProAPO and other text-based baselines by a substantial margin. PEZ benefits from additional shots and improves by 18.7 pp between one and sixteen shots, but it remains clearly worse than other text-based baselines at one shot and only approaches their performance at sixteen shots.

We also compare TPB with the continuous prompt-learning method CoOp~\cite{zhou2022coop}. While CoOp achieves higher accuracy at larger shot counts, this comes at the cost of being coupled to the model's specific representation space. In contrast, TPB preserves the universal nature of language, unlocking robust cross-model transferability where continuous prompts fail as further discussed in \Cref{subsec:exp_transfer}.

\subsubsection{Qualitative Analysis: Targeting Hard Samples.}
\begin{figure}[t]
    \centering
    \begin{subfigure}{0.66\linewidth}
        \centering
        \includegraphics[height=4cm, keepaspectratio]{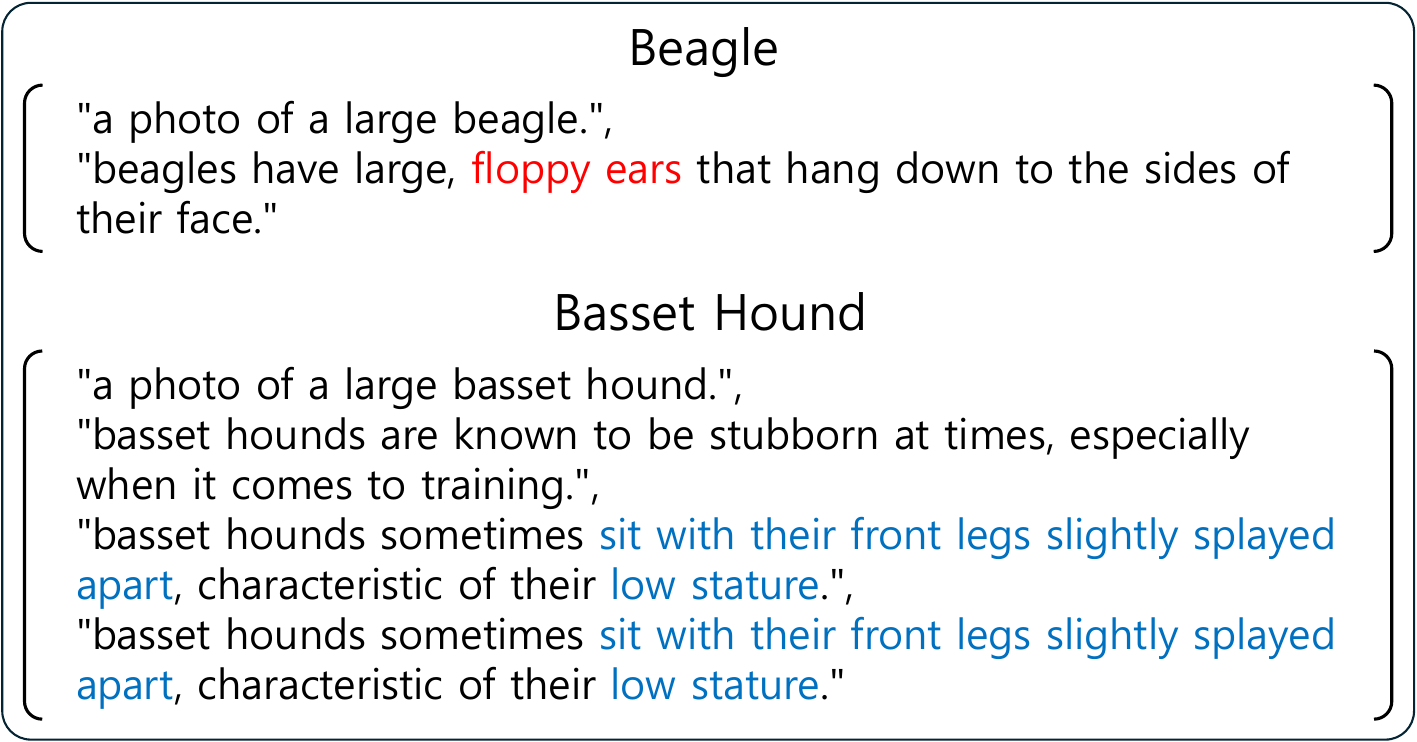}
        \caption{Text Prompts}
        \label{fig:qualitative_example_a}
    \end{subfigure}
    \hfill
    \begin{subfigure}{0.33\linewidth}
        \centering
        \includegraphics[height=4cm, keepaspectratio]{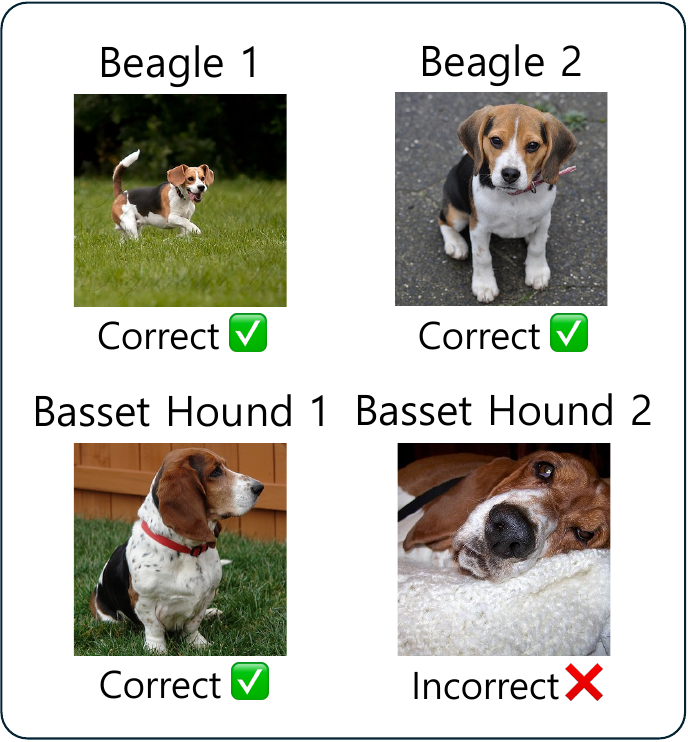}
        \caption{Classification Results}
        \label{fig:qualitative_example_b}
    \end{subfigure}
    
    \caption{
        \textbf{Qualitative example of a weak classifier.} (a) Text prompts for \textit{beagle} and \textit{basset hound}. (b) The weak classifier defined by these prompts correctly classifies three images, but it misclassifies the image of a basset hound lying down, whose ears and legs are mostly occluded.
    }
    \label{fig:qualitative_example}
\end{figure}

To visualize how TPB iteratively builds an ensemble by explicitly targeting misclassified images, we present a qualitative example of the GPC algorithm on a two-class task distinguishing \emph{beagle} from \emph{basset hound}.
As shown in \Cref{fig:qualitative_example_a}, GPC constructs weak classifiers by selecting prompts from the pool that capture dominant, easily recognizable features.
For the \emph{beagle}, it highlights ``large'', ``floppy ears,'' while for the \emph{basset hound}, it emphasizes ``low stature'' and ``front legs slightly splayed apart.''

While this prompt bank correctly classifies standard images as depicted in \Cref{fig:qualitative_example_b}, it fails on a challenging \emph{hard sample}, a basset hound lying down where primary discriminative visual cues like legs and overall stature are heavily occluded.
In a standard text-prompting paradigm, this error would likely remain unresolved.
However, within the TPB framework, this specific misclassification is clearly penalized, assigning the image a higher sample weight in the subsequent boosting round.
Consequently, the next iteration of GPC is explicitly forced to search for alternative textual cues, such as facial wrinkles or distinct snout shapes, specifically to correct this unresolved error.
This highlights the underlying mechanism of TPB: rather than relying on a single static validation objective, the framework adapts to challenging samples by incrementally assembling diverse natural-language prompts.

\subsection{Transfer Robustness Across Heterogeneous VLMs}
\label{subsec:exp_transfer}
\begin{table}[t]
    \caption{
        \textbf{Cross-model transferability of TPB.}
        Average Top-1 accuracy ($\%$) across eleven datasets. All methods are optimized on OpenAI CLIP ViT-B/32 and directly evaluated on larger, heterogeneous target models (ViT-L and ViT-H). TPB preserves shot-driven gains where other text-based baselines saturate, demonstrating superior robustness and scalability during model transfer.
    }
    \label{tab:transfer}

    \centering
    \footnotesize
    \setlength{\tabcolsep}{4pt}
    
    \scalebox{0.9}{
    \begin{tabular}{ll|ccccc}
        \toprule
        \multirow{2}{*}{\textbf{Target Model}} & \multirow{2}{*}{\textbf{Method}} & \multicolumn{5}{c}{\textbf{Shot}} \\
        & & 1 & 2 & 4 & 8 & 16\\
        
        \midrule
        
        \rowcolor{gray!15}
        \multicolumn{7}{c}{\textit{Source Model: OpenAI CLIP, ViT-B/32}}\\
        \midrule
        & ZS-CLIP & \multicolumn{5}{c}{76.76} \\
        \cmidrule(){2-7}
        & CoOp+EFT & 63.81 & 65.98 & 69.45 & 71.78 & 74.05 \\
        & PromptSRC+EFT & 66.79 & 68.07 & 71.65 & 73.45 & 75.20 \\
        \cmidrule(){2-7}
        & PEZ & 49.05 & 53.21 & 55.56 & 57.84 & 58.93 \\
        & ProAPO & \underline{79.59} & \underline{79.53} & \underline{79.87} & \underline{79.93} & \underline{80.01} \\
        \rowcolor{blue!8}
        \cellcolor{white}
        \multirow{-6}{*}{\shortstack[l]{ViT-L: \\ \scriptsize{OpenCLIP},\\ \scriptsize{SigLIP2,} \\ \scriptsize{DFN,} \\ \scriptsize{EVA-02-CLIP,} \\ \scriptsize{CLIPA-v2}}}
        & \textbf{TPB (Ours)} & \textbf{79.66} & \textbf{80.35} & \textbf{81.05} & \textbf{81.88} & \textbf{82.07} \\

        \midrule
        
        & ZS-CLIP & \multicolumn{5}{c}{78.75} \\
        \cmidrule(){2-7}
        & CoOp+EFT & 62.69 & 64.63 & 68.17 & 70.40 & 72.73 \\
        & PromptSRC+EFT & 65.34 & 67.02 & 70.52 & 72.28 & 74.16 \\
        \cmidrule(){2-7}
        & PEZ & 46.97 & 51.50 & 54.59 & 56.47 & 57.60 \\
        & ProAPO & \textbf{82.07} & \underline{81.90} & \underline{82.26} & \underline{82.25} & \underline{82.39} \\
        \rowcolor{blue!8}
        \cellcolor{white}
        \multirow{-6}{*}{\shortstack[l]{ViT-H: \\ \scriptsize{OpenCLIP,}\\ \scriptsize{DFN,}\\ \scriptsize{CLIPA-v2,} \\ \scriptsize{MetaCLIP2}}}
        & \textbf{TPB (Ours)} & \underline{81.73} & \textbf{82.12} & \textbf{83.24} & \textbf{83.84} & \textbf{84.24} \\

        \bottomrule
    \end{tabular}
    }
\end{table}

\subsubsection{Source and Target VLMs.}
In our cross-model transfer experiments, we distinguish between `source models' (used for prompt optimization) and `target models' (used for evaluation). Note that the concept of a source model is not applicable to zero-shot methods, as they do not involve model-specific optimization. As sources, we use OpenAI CLIP~\cite{radford2021learning} with two backbones: ResNet-50 and ViT-B/32. For target models, we use OpenCLIP~\cite{cherti2023openclip}, EVA-02-CLIP~\cite{sun2023evaclip}, CLIPA-v2~\cite{li2023clipa}, DFN~\cite{fang2023dfn}, SigLIP2~\cite{tschannen2025siglip2}, and MetaCLIP2~\cite{chuang2025metaclip2}. These VLMs are released with various backbones, which we group by their size (\eg, ViT-L, ViT-H) and report averaged results for each group. This setup covers a range of architectures, scales, and training recipes, allowing us to assess model-transfer performance across heterogeneous VLMs.

\subsubsection{Transfer Baselines.}
For model-transfer experiments, we evaluate both the text-based methods and prompt-learning baselines. As a representative prompt-learning method, we use PromptSRC~\cite{khattak2023promptsrc}, which adapts VLMs by learning visual and textual prompt tokens while keeping the backbone parameters frozen. Since these learned vectors cannot be directly reused in other VLMs with different architectures or representation spaces, we additionally employ emulated fine-tuning (EFT)~\cite{mitchell2023eft} proposed in the language domain. EFT combines a large pretrained model with a smaller fine-tuned model by reweighting logits, effectively emulating the result of fine-tuning the large model without updating its parameters. This allows us to transfer the adaptation learned by PromptSRC to other VLMs in a training-free manner, but the inference overhead increases due to multiple forward passes (see the supplementary material for details).

\subsubsection{Transfer-robust Shot Scalability.}
\Cref{tab:transfer} reports cross-model transfer results when all methods operate on OpenAI CLIP ViT-B/32 with one to sixteen shot supervision and then evaluated on larger target models (ViT-L and ViT-H). On ViT-L targets, TPB improves the average accuracy from 76.76\% (zero-shot CLIP) to 82.07\%; on ViT-H targets, it improves from 78.75\% to 84.24\%. TPB also outperforms ProAPO, the strongest text-based baseline, by 2.06 pp on ViT-L targets and 1.85 pp on ViT-H targets.
Notably, TPB also maintains a performance margin over CoOp+EFT and PromptSRC+EFT. While continuous methods like CoOp can achieve higher accuracy on the source model by overfitting to its specific representation space, they fail to preserve this advantage during transfer.
In contrast, TPB achieves superior transferability and scalability while requiring only a single forward pass during inference, whereas EFT-based transfer necessitates multiple inference passes.
Moreover, TPB exhibits superior transfer-robust shot scalability. While ProAPO’s performance gains from one-shot to sixteen-shot are limited to 0.42 pp on ViT-L and 0.32 pp on ViT-H, TPB achieves significantly higher improvements of 2.41 pp and 2.51 pp, respectively. This contrast demonstrates that TPB consistently preserves shot-driven gains where other text-based baselines saturate, capturing architecture-agnostic task knowledge that remains robust during cross-model transfer.
Conversely, PEZ, which optimizes discrete prompts via gradients, transfers poorly and falls even below zero-shot performance, as we discuss in the following analysis.

Additional shot-wise transfer results on larger target VLMs (\eg, ViT-E and ViT-G) are provided in the supplementary material.
Consistent with \Cref{tab:transfer}, they show that TPB preserves shot-driven gains after transfer.

\subsubsection{Failure Analysis of Gradient-optimized Prompts.}
\begin{table}[t]
    \caption{
        \textbf{Effect of gradient-optimized text prompts on model transfer.}
        Transfer accuracy (\%) over eleven datasets, adapted on RN50 and transferred to ViT-L/14.
    }
    \label{tab:pez}
    
    \centering
    \footnotesize
    \setlength{\tabcolsep}{4pt}

    \scalebox{0.8}{
    \begin{tabular}{c|ccc}
    \toprule
    \multirow{2}{*}{\textbf{Backbone}} & \multicolumn{3}{c}{\textbf{Methods}} \\
     & ZS-CLIP & PEZ & TPB (ours) \\
    \midrule
    RN50 (source) & 55.98 & \acc{\underline{61.09}}{0.79} & \acc{\mathbf{70.32}}{0.51} \\
    ViT-L/14 (target) & \underline{71.07} & \acc{53.02}{1.75} & \acc{\mathbf{78.46}}{0.36} \\
    \bottomrule
    \end{tabular}
    }
\end{table}

\Cref{tab:pez} reports the six\-teen-shot average accuracy over eleven
datasets when adapting OpenAI CLIP RN50 and transferring the resulting
prompts to OpenAI CLIP ViT-L/14. On the source model, PEZ improves
zero-shot CLIP from $55.98\%$ to $61.09\%$, indicating that it
successfully adapts the RN50 backbone to the training tasks. However, the
learned prompts transfer poorly: on the ViT-L/14 target, accuracy drops
from $71.07\%$ (zero-shot CLIP) to $53.02\%$, well below even the
zero-shot baseline.
To understand why PEZ transfers so poorly despite using text prompts, we
inspect the optimized context tokens. For example, for the \textit{pug}
class in OxfordPets, PEZ yields the following prompts:
\begin{quote}
    \centering
    ``hey darby pls violets kissestc accept liza any,, \\adorable gorgeous ♡♡ :-) life behaved \textit{pug}''
\end{quote}
Here, every token except the final class name ``\textit{pug}'' is obtained
by gradient updates during optimization. The resulting context is a sequence of seemingly arbitrary words, symbols, and non-word fragments with no coherent natural-language meaning. Such prompts behave like model-specific codes in the RN50 representation space, rather than semantically meaningful descriptions that a human could interpret.

This observation suggests that the ability of a prompt to transfer across
models does not come simply from it being text. Instead, prompts are more
transferable when they stay close to fluent, human-readable natural-language. Prompts that drift too far from human-interpretable
language may overfit a particular model and therefore fail to transfer.

\subsection{Ablation Studies}
\subsubsection{Effect of Augmentation and Prompt Pool Diversity.}
{
    \pgfplotsset{compat=1.18}
    \pgfplotsset{
        shotaxis/.style={
            label style={font=\scriptsize},
            tick label style={font=\scriptsize},
            width=1.32\linewidth,
            height=1.2\linewidth,
            xlabel=Shot,
            ylabel=Acc.(\%),
            xmin=-0.3, xmax=17.3,
            ymin=63.1, ymax=73.9,
            xlabel style={yshift=0.10cm},
            ylabel style={yshift=-0.15cm},
            xtick={1,2,4,8,16},
            ytick={64, 66, 68, 70, 72, 74},
            legend style={
                at={(1,0)},
                anchor=south east,
                draw=none,
                fill=none,
                font=\tiny,
                legend cell align=left,
                row sep=-2pt,
            },
            legend columns=1,
            legend image code/.code={
                \draw[#1] (0cm,0cm) -- (0.34cm,0cm);
                \draw[mark=halfcircle*, mark options={fill=white, draw=#1}]
                plot coordinates {(0.17cm,0cm)};
                \draw[#1] plot coordinates {(0.17cm,0cm)};
            }
        }
    }
    
    \begin{figure}[t]
        \centering
        \pgfplotsset{every axis/.append style={mark size=1.4pt}}
    
        \hfill
        \hfill
        \hfill
        \hfill
        \begin{subfigure}{.3\linewidth}
            \centering
            \begin{tikzpicture}[trim axis left, trim axis right]
                \begin{axis}[shotaxis]
                    \addplot[cRed, semithick, mark=*, mark options={solid}]
                        coordinates {(-1, -1)};
                    \addplot[cOrange, semithick, mark=halfcircle*, mark options={solid, rotate=240}]
                        coordinates {(-1, -1)};
                    \addplot[cYellow, semithick, mark=halfcircle, mark options={solid, rotate=60}]
                        coordinates {(-1, -1)};
                    \addplot[Goldenrod,   semithick, mark=*, mark options={solid, fill=white}]
                        coordinates {(-1, -1)};
              
                    \addplot[Goldenrod,   semithick, mark=*, mark options={solid, fill=white}]
                        table[col sep=comma, x=x, y=x0]{figure/fig_ablation.csv};
                    \addplot[cYellow,  semithick, mark=*, mark options={solid, fill=white}]
                        table[col sep=comma, x=x, y=x1]{figure/fig_ablation.csv};
                    \addplot[cYellow,  semithick, only marks, mark=halfcircle, mark options={solid, rotate=60}]
                        table[col sep=comma, x=x, y=x1]{figure/fig_ablation.csv};
                    \addplot[cOrange, semithick, mark=halfcircle*, mark options={solid, rotate=240}]
                        table[col sep=comma, x=x, y=x2]{figure/fig_ablation.csv};
                    \addplot[cRed,    semithick, mark=*, mark options={solid}]
                        table[col sep=comma, x=x, y=x4]{figure/fig_ablation.csv};
                    \legend{$a$=4, $a$=2, $a$=1, No aug}
                \end{axis}
            \end{tikzpicture}
            \caption{Augmentation factor $a$}
            \label{fig:ablation_augmentation}
        \end{subfigure}
        \hfill
        \begin{subfigure}{.3\linewidth}
        \centering
            \begin{tikzpicture}[trim axis left, trim axis right]
                \begin{axis}[shotaxis]
                    \addplot[cRed, semithick, mark=*, mark options={solid}]
                        coordinates {(-1, -1)};
                    \addplot[Orchid, semithick, mark=halfcircle*, mark options={solid, rotate=240}]
                        coordinates {(-1, -1)};
                    \addplot[Thistle,   semithick, mark=*, mark options={solid, fill=white}]
                        coordinates {(-1, -1)};
            
                    \addplot[Thistle,   semithick, mark=*, mark options={solid, fill=white}]
                        table[col sep=comma, x=x, y=cupl]{figure/fig_ablation.csv};
                    \addplot[Orchid, semithick, mark=halfcircle*, mark options={solid, rotate=240}]
                        table[col sep=comma, x=x, y=cupldclip]{figure/fig_ablation.csv};
                    \addplot[cRed,    semithick, mark=*, mark options={solid}]
                        table[col sep=comma, x=x, y=all]{figure/fig_ablation.csv};
                    \legend{All descriptions, CuPL+DCLIP, CuPL}
                \end{axis}
            \end{tikzpicture}
            \caption{Prompt pool ablation}
            \label{fig:ablation_bank}
        \end{subfigure}
        \hfill
        \hfill
        \hfill
        \hfill
        
        \caption{
            \textbf{Ablation of each component of TPB.}
            Average accuracy over eleven datasets when using OpenAI CLIP ViT-B/32 as both source and target model.
            (a) Performance for different augmentation factors $a$
            (b) Performance for different constructions of the prompt pool.
        }
        \label{fig:ablation}
    \end{figure}
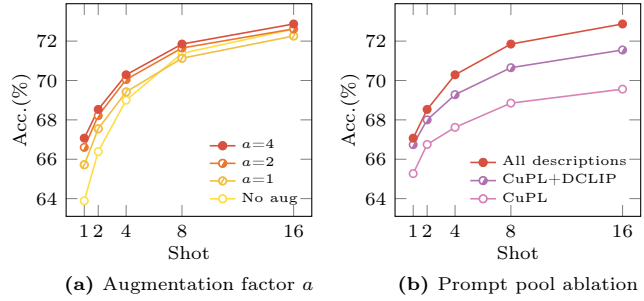
}

In \Cref{fig:ablation_augmentation}, increasing the augmentation factor $a$ consistently improves accuracy over using no augmentation, with the largest gains in the low-shot setting.
Here, $a$ controls the number of random transformations per training example, and $a=0$ corresponds to using only the fixed few-shot set without any augmentation.
Without augmentation, AdaBoost quickly fits the small training set within a few rounds, leaving little signal for subsequent boosting rounds, whereas randomly transformed views of the same images keep introducing informative errors, enabling later weak classifiers to correct them and yielding a stronger ensemble.
\Cref{fig:ablation_bank} then examines the impact of the diversity of the prompt pool, constructed by combining several groups of LLM-generated descriptions.
Using both CuPL and DCLIP descriptions yields better performance than using CuPL alone, even though DCLIP is individually weaker, as reflected by its lower accuracy in \Cref{tab:mainshot}.
Incorporating all available descriptions further improves accuracy, especially at higher shot counts, suggesting that a more diverse prompt pool helps TPB better exploit additional labeled examples as they become available.

\subsubsection{Dynamics of Augmentation Across Boosting Rounds.}
{
    \pgfplotsset{compat=1.18}
    \pgfplotsset{
        shotaxis/.style={
            label style={font=\scriptsize},
            tick label style={font=\scriptsize},
            width=1.32\linewidth,
            height=1.2\linewidth,
            xlabel=Boosting round,
            ylabel=Acc. (\%),
            xmin=-2, xmax=53,
            xtick={1, 10, 30, 50},
            ytick={55, 65, 75, 85, 95},
            xlabel style={yshift=0.15cm},
            ylabel style={yshift=-0.15cm},
            legend style={
                at={(1,0.4)},
                anchor=east,
                draw=none,
                fill=none,
                font=\tiny,
                legend cell align=left,
                row sep=-2pt,
            },
            legend columns=1,
            legend image code/.code={
                \draw[#1] (0cm,0cm) -- (0.34cm,0cm);
                \draw[mark=halfcircle*, mark options={fill=white, draw=#1}]
                plot coordinates {(0.17cm,0cm)};
                \draw[#1] plot coordinates {(0.17cm,0cm)};
            }
        }
    }
    
    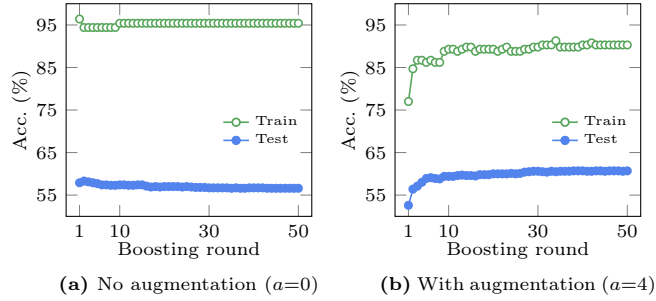
\begin{figure}[t!]
        \centering
        \pgfplotsset{every axis/.append style={mark size=1.4pt}}

        \hfill
        \hfill
        \hfill
        \hfill
        \begin{subfigure}{.3\linewidth}
            \centering
            \begin{tikzpicture}[trim axis left, trim axis right]
                \begin{axis}[shotaxis, ymin=50, ymax=100]
                    \addplot[cGreen,   semithick, mark=*, mark options={solid, fill=white}, restrict x to domain=0:50]
                        table[col sep=comma, x expr=\thisrow{x}+1, y expr=\thisrow{train_acc}*100]{figure/fig_stanfordcars_noaug.csv};
                    \addplot[cBlue, semithick, mark=*, mark options={solid}, restrict x to domain=0:50]
                        table[col sep=comma, x expr=\thisrow{x}+1, y expr=\thisrow{test_acc}*100]{figure/fig_stanfordcars_noaug.csv};
                    \legend{Train, Test}
                \end{axis}
            \end{tikzpicture}
            \caption{No augmentation ($a{=}0$)}
            \label{fig:augmenttesta}
        \end{subfigure}
        \hfill
        \begin{subfigure}{.3\linewidth}
            \centering
            \begin{tikzpicture}[trim axis left, trim axis right]
                \begin{axis}[shotaxis, ymin=50, ymax=100]
                    \addplot[cGreen,   semithick, mark=*, mark options={solid, fill=white}, restrict x to domain=0:50]
                        table[col sep=comma, x expr=\thisrow{x}+1, y expr=\thisrow{train_acc}*100]{figure/fig_stanfordcars_4aug.csv};
                    \addplot[cBlue, semithick, mark=*, mark options={solid}, restrict x to domain=0:50]
                        table[col sep=comma, x expr=\thisrow{x}+1, y expr=\thisrow{test_acc}*100]{figure/fig_stanfordcars_4aug.csv};
                    \legend{Train, Test}
                \end{axis}
            \end{tikzpicture}
            \caption{With augmentation ($a{=}4$)}
            \label{fig:augmenttestb}
        \end{subfigure}
        \hfill
        \hfill
        \hfill
        \hfill
    
    \caption{\textbf{Effects of augmentation over boosting rounds.}
    Train and test accuracy on StanfordCars in the one-shot setting using OpenAI CLIP ViT-B/32.}
    \label{fig:augmenttest}
    
    \end{figure}
}

\Cref{fig:augmenttest} compares AdaBoost with and without image augmentation.
Without augmentation (\Cref{fig:augmenttesta}), training the ensemble rapidly memorizes the few available examples, pushing training accuracy close to $100\%$ within only a few rounds.
Once training accuracy saturates, the reweighted distribution no longer highlights informative mistakes, preventing AdaBoost from meaningfully updating its weak learners; consequently, test accuracy remains almost unchanged across subsequent rounds.
In contrast, when applying image augmentation with a factor of $a{=}4$ (\Cref{fig:augmenttestb}), each boosting round observes newly transformed views of the same images.
These variations prevent the model from perfectly fitting the training set, keeping training accuracy below saturation and ensuring that each round still produces informative errors for reweighting.
This continual supply of hard examples stabilizes the boosting process and allows test accuracy to improve steadily over many rounds, demonstrating that augmentation is essential for effective boosting under low-shot supervision.

\subsubsection{Effect of the Number of Weak Classifiers.}
\begin{table}[t]
    \caption{\textbf{Ablation on the number of weak classifiers.} Results (\%) averaged over 10 datasets excluding ImageNet-1K using OpenAI CLIP ViT-B/32.}
    \label{tab:weaknumber}

    \centering
    \footnotesize
    \setlength{\tabcolsep}{4pt}

    \scalebox{0.8}{
    \begin{tabular}{c|c|cccc}
        \toprule
        \multirow{2}{*}{\textbf{Shots}}
        & \multirow{2}{*}{\textbf{ProAPO}} & \multicolumn{4}{c}{\textbf{TPB (ours)}} \\
        &  & \textbf{$M=1$} & \textbf{$M=10$} & \textbf{$M=30$} & \textbf{$M=50$} \\
        \midrule
        1  & \acc{66.27}{0.46} & \acc{61.80}{0.74} & \acc{66.77}{0.44} & \acc{\underline{67.28}}{0.77} & \acc{\mathbf{67.39}}{0.81} \\
        16 & \acc{67.74}{0.39} & \acc{70.13}{0.42} & \acc{72.84}{0.49} & \acc{\underline{73.21}}{0.44} & \acc{\mathbf{73.36}}{0.47} \\
        \bottomrule
    \end{tabular}
    }
\end{table}

\Cref{tab:weaknumber} analyzes how the performance of TPB varies with the number of weak classifiers $M$.
With only a single weak classifier ($M{=}1$), TPB is slightly weaker than ProAPO in the one-shot setting.
However, once $M$ reaches $10$, TPB already matches or exceeds ProAPO at both one-shot and sixteen-shot.
Increasing $M$ further to $30$ or $50$ provides additional but diminishing improvements, suggesting that performance effectively saturates in the range of $M{=}30$-$50$.

\subsubsection{Boosting rounds \vs. augmentation.}

\begin{table}[t!]
    \caption{\textbf{Boosting \vs. augmentation.} Top-1 accuracy (\%) in the 16-shot setting, averaged over 10 datasets excluding ImageNet with fixed exposure $M\!\cdot\!a=200$.}
    \label{tab:ma}

    \centering
    \footnotesize
    \setlength{\tabcolsep}{4pt}

    \scalebox{0.8}{
    \begin{tabular}{cc|c|cccc}
        \toprule
        \multicolumn{2}{c|}{\multirow{2}{*}{Backbones}} & \multirow{2}{*}{ZS-CLIP} & \multicolumn{4}{c}{$M/a$}\\
        & & & $1/200$ & $5/40$ & $10/20$ & $50/4$ \\
        \midrule
        Source & ViT-B/32 & 60.10 & \acc{\mathbf{74.42}}{0.17} & \acc{72.94}{0.10} & \acc{73.32}{0.26} & \acc{\underline{73.36}}{0.47} \\
        \midrule
        \multirow{2}{*}{Target} & ViT-L & 76.53 & \acc{81.54}{0.13} & \acc{81.78}{0.13} & \acc{\underline{82.13}}{0.03} & \acc{\mathbf{82.24}}{0.39} \\
        & ViT-H & 78.60 & \acc{83.12}{0.26} & \acc{83.72}{0.19} & \acc{\underline{83.98}}{0.25} & \acc{\mathbf{84.49}}{0.53} \\
        \bottomrule
    \end{tabular}
    }
\end{table}

To isolate boosting from image augmentation, we fix the total exposure budget $M{\cdot}a{=}200$ and sweep $(M, a)$ as shown in~\Cref{tab:ma}.
This sweep is conducted in the 16-shot setting over ten datasets excluding ImageNet, using OpenAI CLIP ViT-B/32 as the source model and averaging transfer accuracy over the same target groups as in~\Cref{tab:transfer}: five ViT-L-scale and four ViT-H-scale VLMs.
Under this fixed exposure budget, increasing $a$ while reducing $M$ improves source accuracy but degrades transfer accuracy.
This comparison shows that transfer robustness is driven by iterative boosting rather than by augmented exposure alone.

\section{Conclusion}
We introduce Text Prompt Boosting (TPB), a boosting-based framework that ensembles natural-language prompts for shot-scalable and transfer-robust few-shot adaptation of vision-language models.
By treating a class-wise prompt bank collection as a weak classifier and employing the Greedy Prompt Composition (GPC) procedure as the weak learner, TPB incrementally builds an ensemble of prompt-based classifiers that explicitly focus on hard examples while preserving interpretability and transferability.
Consequently, the resulting strong classifier exhibits improved shot scalability on the source model, while simultaneously achieving transfer robustness, successfully preserving its shot-driven performance gains even when evaluated on various larger heterogeneous VLMs.

\subsubsection{Limitations:}
Despite its effectiveness, TPB has limitations.
First, its construction cost can be non-negligible on large-scale datasets, where the candidate space is substantially larger.
Second, AdaBoost-style reweighting may overemphasize noisy or atypical few-shot samples.
Finally, TPB selects prompts from a fixed pool, which may limit its flexibility for fine-grained or ambiguous classes.

\section*{Acknowledgements}
This work was supported by the Institute of Information \& Communications Technology Planning \& Evaluation (IITP) grant funded by the Korea government (MSIT) (Nos. RS-2019-II191906, Artificial Intelligence Graduate School Program; RS-2024-00457882, AI Research Hub Project; RS-2026-25511821, Development of Personalized Media Service Recommendation and Generative Technology), by a grant (RS-2025-00564342) from the Korea Institute for Advancement of Technology (KIAT), funded by the Ministry of Trade, Industry and Energy (MOTIE), and by Seoul R\&D Program (SP240008) through the Seoul Business Agency (SBA) funded by the Seoul Metropolitan Government.

%
%
\bibliographystyle{splncs04}
\bibliography{main}
\clearpage

\appendix

\section*{Supplementary Materials}
This supplementary material provides (A) further details of the Greedy Prompt Composition (GPC); (B) a summary of the SAMME.R algorithm used in our boosting framework; (C) experimental settings, including hardware setup and baseline implementation details; and (D) additional experiments and analyses.

\section{Methodology Details}
\label{sup:sec:method}

\begin{algorithm}[h]
    \caption{Our weak learning algorithm, Greedy Prompt Composition (GPC). In Line 9, \textsc{SelectBest} denotes prompt selection process, which is detailed in \Cref{sec:selectbest}.}
    \label{alg:gpc}
    \begin{algorithmic}[1]
        \Require Labeled dataset $\mathcal{D}=\{(\mathbf{x}_i,y_i)\}_{i=1}^n$ with $y_i\in\{1,\dots,K\}$; sample weights $\mathbf{w}=(w_1,\dots,w_n)$; template bank $\Phi_\mathrm{template}$; large prompt pool $\mathcal{P} = \{P_1, \dots, P_K\}$.
        \State \textbf{Stage 1: Single-Template Initialization}
        \State Select the template $\phi^\star \in \Phi_\mathrm{template}$ that minimizes the weighted error.
        \State Initialize class-wise prompt banks $B_k \leftarrow [\phi^\star(c_k)]$ for all $k \in \{1,\dots,K\}$, where $\phi^\star(c_k)$ denotes $\phi^\star$ with the name of class $k$ inserted.
        \State \textbf{Stage 2: Class-wise Greedy Prompt Insertion}
        \While{true}
            \State $\textit{improved} \leftarrow \mathrm{false}$
            \For{$k = 1,2,\dots,K$}
                \State $\mathcal{B} \leftarrow \{B_1,\dots,B_K\}$
                \State $(t, d,\Delta\varepsilon) \leftarrow \textsc{SelectBest}(\mathcal{D}, \mathbf{w}, \mathcal{P}, \mathcal{B}, k)$ \Comment{See \Cref{sec:selectbest}}
                \If{$\Delta\varepsilon < 0$}
                    \State $B_k \leftarrow B_k^{(t,d)}$ \Comment{Append $d$ copies of $t$ to $B_k$}
                    \State $\textit{improved} \leftarrow \mathrm{true}$
                \EndIf
            \EndFor
            \If{not $\textit{improved}$}
                \State \textbf{break}
            \EndIf
        \EndWhile
        \State $\mathcal{B}^\star \leftarrow \{B_1,\dots,B_K\}$
        \State \textbf{Output:} weak classifier $F$ defined by $\mathcal{B}^\star$:
        \[
            F(\mathbf{x})
                =
            \Big( p(y=1 \mid \mathbf{x}; \mathcal{B}^\star),\; \dots,\; p(y=K \mid \mathbf{x}; \mathcal{B}^\star) \Big),
        \]
        where $p(y=k \mid \mathbf{x}; \mathcal{B^\star})$ is defined in \Cref{eq:prob_scalar}.
    \end{algorithmic}
\end{algorithm}

In \Cref{subsec:gpc}, we introduce Greedy Prompt Composition (GPC) as the weak learning algorithm used in our AdaBoost framework.
At boosting round $m$, GPC constructs a weak classifier by composing class-specific prompt banks $B_k^{(m)}$ drawn from the large prompt pool $P_k$ for each class $k \in \{1,\dots,K\}$. The procedure consists of two stages: (i) single-template initialization and (ii) class-wise greedy prompt insertion, which repeatedly selects the prompt that most reduces the weighted error in \Cref{eq:weighted_error}. \Cref{alg:gpc} summarizes the overall flow. The remainder of this section details Stage~2, \ie, the class-wise greedy prompt insertion.

\subsection{Prompt Duplication}
Given the current prompt-bank collection $\mathcal{B} = \{B_1, \dots, B_K\}$ and a target class $k$, each iteration of Stage 2 selects a prompt $t^\star$ from the pool $P_k$ and appends $d \in \mathbb{N}$ copies of it to the current bank $B_k$.
Let $n_k = |B_k|$ denote the current size of $B_k$, and let $B_k^{(t,d)}$ denote the bank obtained by adding $d$ copies of a candidate prompt $t \in P_k$ to $B_k$.
Since the class score $s(\mathbf{x}, B_k)$ is defined as the unweighted average similarity over the prompts in a bank, adding $d$ copies of $t$ changes the score for class $k$ on an input $\mathbf{x}$ to:
\begin{equation}
    s(\mathbf{x}, B_k^{(t,d)})
        =
    \frac{n_k\,s(\mathbf{x}, B_k) + d\,s(\mathbf{x}, t)}{n_k + d}.
    \label{eq:score-update}
\end{equation}
While we can set $d=1$ (no duplication), adding only a single copy to an already populated bank often yields a negligible change in the class score and fails to improve the weighted error.
To better capture the influence of each candidate, we simply allow for duplicate selections of the same prompt.
This approach serves as a discrete approximation of continuous weighting, allowing us to adjust the effective contribution of a prompt without introducing learnable weights or gradient updates.

\subsection{Prompt Selection with Duplication}
\label{sec:selectbest}
We now derive a closed-form expression for the change in weighted error when $d$ copies of a candidate prompt $t\in P_k$ are appended to $B_k$.
Let $y_i$ denote the ground-truth label of $\mathbf{x}_i$, $\hat y_i$ denote the prediction of the current classifier, and let $\hat y_i^{\mathrm{2nd}}$ denote the class with the second largest class score:
\begin{equation}
    \begin{aligned}
        \hat y_i &:= \argmax_{j \in \{1,\dots,K\}} s(\mathbf{x}_i, B_j)\;, \\
        \hat y_i^{\mathrm{2nd}} &:= \argmax_{j \in \{1,\dots,K\} \setminus \{\hat y_i\}} s(\mathbf{x}_i, B_j)\;.        
    \end{aligned}
\end{equation}
Since modifying $B_k$ only changes the score of class $k$, the prediction for each $\mathbf{x}_i$ can change in only three ways: (i) no change, (ii) an \textbf{upside flip} ($k$ overtakes $\hat y_i$), or (iii) a \textbf{downside flip} ($k = \hat y_i$ and $k$ is overtaken by $\hat y_i^{\mathrm{2nd}}$).

\subsubsection{Upside flips.}
An upside flip for sample $i$ occurs when $\hat y_i \neq k$ and the updated score of class $k$ overtakes that of the current winner:
\begin{equation}
    \begin{aligned}
        s(\mathbf{x}_i, B_k^{(t,d)})
            &= \frac{n_k s(\mathbf{x}_i, B_k) + d\,s(\mathbf{x}_i, t)}{n_k + d}
            > s(\mathbf{x}_i, B_{\hat y_i}) \\
        &\iff
        \frac{n_k\big(s(\mathbf{x}_i, B_k)-s(\mathbf{x}_i, B_{\hat y_i})\big)}
             {s(\mathbf{x}_i, B_{\hat y_i})-s(\mathbf{x}_i,t)} < d,
    \end{aligned}
\end{equation}
which is meaningful only when $s(\mathbf{x}_i, B_{\hat y_i}) < s(\mathbf{x}_i, t)$; otherwise, no finite $d$ can make class $k$ overtake $\hat y_i$. The minimum integer $d$ that causes an upside flip for $\mathbf{x}_i$ is therefore
\begin{equation}
    d_i^{\uparrow}(t) :=
    \begin{cases}
        \begin{aligned}
            &\left\lfloor \dfrac{n_k\big(s(\mathbf{x}_i, B_k)-s(\mathbf{x}_i, B_{\hat y_i})\big)}{s(\mathbf{x}_i, B_{\hat y_i})-s(\mathbf{x}_i,t)}\right\rfloor + 1\;, \\
            &\qquad \text{if } \hat y_i\neq k \text{ and } s(\mathbf{x}_i,B_{\hat y_i}) < s(\mathbf{x}_i, t)\;,
        \end{aligned} \\\\
        \begin{aligned}
            &+\infty\;, \\
            &\qquad \text{otherwise\;,}
        \end{aligned}
    \end{cases}
\end{equation}

where $+\infty$ indicates that upside flips for sample $i$ never occur, so that $\mathbb{I}[d \ge d_i^{\uparrow}(t)] = 0$ for all finite $d$.

Upside flips are beneficial when they correct a misclassified sample (\ie, when $y_i = k$), and harmful when they change a previously correct prediction into an incorrect one (\ie, when $y_i = \hat y_i$ and $y_i \neq k$). We therefore define the potentially beneficial and harmful index sets for upside flips as
\[
    \mathcal{B}^\uparrow_k := \{i : y_i = k\},
    \qquad
    \mathcal{H}^\uparrow_k := \{i : y_i = \hat y_i,\; y_i \neq k\}.
\]

\subsubsection{Downside flips.}
A downside flip for sample $i$ occurs when $\hat y_i = k$ and the updated score of class $k$ falls below that of the second-best class:
\begin{equation}
    \begin{aligned}
        s(\mathbf{x}_i, B_k^{(t,d)})
            &= \frac{n_k s(\mathbf{x}_i, B_k) + d\,s(\mathbf{x}_i, t)}{n_k + d}
            < s(\mathbf{x}_i, B_{\hat y_i^\mathrm{2nd}}) \\
        &\iff
        \frac{n_k\big(s(\mathbf{x}_i, B_k)-s(\mathbf{x}_i, B_{\hat y_i^\mathrm{2nd}})\big)}
             {s(\mathbf{x}_i, B_{\hat y_i^\mathrm{2nd}})-s(\mathbf{x}_i,t)} < d,
    \end{aligned}
\end{equation}
which is meaningful only when $s(\mathbf{x}_i, t) < s(\mathbf{x}_i,B_{\hat y_i^\mathrm{2nd}})$. The minimum integer $d$ that causes a downside flip for $\mathbf{x}_i$ is
\begin{equation}
    d_i^{\downarrow}(t) :=
    \begin{cases}
        \begin{aligned}
            &\left\lfloor \dfrac{n_k\big(s(\mathbf{x}_i, B_k)-s(\mathbf{x}_i, B_{\hat y_i^\mathrm{2nd}})\big)}{s(\mathbf{x}_i, B_{\hat y_i^\mathrm{2nd}})-s(\mathbf{x}_i,t)}\right\rfloor + 1\;, \\
            &\qquad \text{if } \hat y_i= k \text{ and } s(\mathbf{x}_i, t) < s(\mathbf{x}_i,B_{\hat y_i^\mathrm{2nd}})\;,
        \end{aligned} \\\\
        \begin{aligned}
            &+\infty\;, \\
            &\qquad \text{otherwise\;.}
        \end{aligned}
    \end{cases}
\end{equation}
Downside flips are beneficial when they fix a previously incorrect prediction (\ie, when the new prediction $\hat y_i^\mathrm{2nd}$ equals $y_i$) and harmful when they corrupt a previously correct prediction (\ie, when $y_i = k$). Accordingly, we define
\[
    \mathcal{B}^\downarrow_k := \{i : y_i = \hat y_i^\mathrm{2nd}\},
    \qquad
    \mathcal{H}^\downarrow_k := \{i : y_i = k\}.
\]

\subsubsection{Change in weighted error.}
Let $\varepsilon(\mathcal{B})$ denote the weighted error in \Cref{eq:weighted_error} of the classifier defined by the prompt collection $\mathcal{B}$, and let $\mathcal{B}^{(k,t,d)}$ denote the collection obtained by replacing $B_k$ with $B_k^{(t,d)}$.
The change in weighted error when adding $d$ copies of $t$ to $B_k$ is
\begin{equation}
    \Delta \varepsilon(k,t,d) := \varepsilon\big(\mathcal{B}^{(k,t,d)}\big) - \varepsilon(\mathcal{B})\;. 
\end{equation}
Recall that adding copies of $t$ only modifies the score of class $k$. Thus, as we vary the candidate prompt $t$ and the duplication factor $d$, the only way the weighted error can change is when the prediction for some training example flips to a different class. As described above, every such change in prediction is either an upside flip (from $\hat y_i$ to $k$) or a downside flip (from $k$ to $\hat y_i^\mathrm{2nd}$). Consequently, the change in weighted error can be decomposed into contributions from these upside and downside flips:
\begin{equation}
    \begin{aligned}
        \Delta \varepsilon(k,t,d)
        &= \sum_{i\in \mathcal{H}^\uparrow_k} w_i\,\mathbb{I}\big[d\geq d^\uparrow_i(t)\big]
            +  \sum_{i\in \mathcal{H}^\downarrow_k} w_i\,\mathbb{I}\big[d\geq d^\downarrow_i(t)\big] \\
        &- \sum_{i\in \mathcal{B}^\uparrow_k} w_i\,\mathbb{I}\big[d\geq d^\uparrow_i(t)\big]
            - \sum_{i\in \mathcal{B}^\downarrow_k} w_i\,\mathbb{I}\big[d\geq d^\downarrow_i(t)\big]\;.
    \end{aligned}
    \label{eq:delta-eps}
\end{equation}
The first two sums aggregate harmful flips ($i\in\mathcal H^\uparrow_k \cup \mathcal H^\downarrow_k$), which increase the weighted error once their indicators become 1. The last two sums aggregate beneficial flips ($i\in\mathcal B^\uparrow_k \cup \mathcal B^\downarrow_k$) with a minus sign, so they decrease the weighted error when they occur.
As a function of $d$, $\Delta \varepsilon(k,t,d)$ is a right-continuous step function whose value can change only at the (finite) integer thresholds where upside or downside flips occur:
\begin{equation}
    \begin{aligned}
        \mathcal{T}(t)
        :=
        &\big\{d^\uparrow_i(t) \mid 1\le i \le n,\ d^\uparrow_i(t)<\infty\big\}\\
        &\cup \big\{d^\downarrow_i(t) \mid 1\le i \le n,\ d^\downarrow_i(t)<\infty\big\}\;.        
    \end{aligned}
\end{equation}
Hence we can restrict our search for the optimal number of copies $d$ to $\mathcal{T}(t)$ instead of the infinite set $\mathbb{N}$.

We choose the optimal number of copies for candidate prompt $t$ as
\begin{equation}
    d(t) = \argmin_{d \in \mathcal{T}(t)} \frac{\Delta \varepsilon(k,t,d)}{d}\;,
    \label{eq:dup-opt}
\end{equation}
where ties are broken in favor of smaller $d$; see \Cref{sec:dup-penalty} for motivation.
Using this, the best prompt for class $k$ is selected as
\begin{equation}
    t^\star
    =
    \argmin_{t\in P_k}
    \frac{\Delta \varepsilon\big(k,t,d(t)\big)}{d(t)}.
    \label{eq:best-prompt}
\end{equation}
If the corresponding error change is negative, \ie,
$\Delta \varepsilon\big(k,t^\star,d(t^\star)\big) < 0$, we update the current prompt bank as
$B_k \leftarrow B_k^{(t^\star, d(t^\star))}$ and proceed to the next class in \Cref{alg:gpc}.

\subsection{Penalty on Larger Duplication}
\label{sec:dup-penalty}
Prior work on text prompts for VLMs suggests that using a diverse set of prompts for each class improves robustness and generalization~\cite{menon2022dclip,roth2023waffling,pratt2023cupl}. Because class scores are defined as averages over prompts in $B_k$, prompts whose similarity $s(\mathbf{x},t)$ is close to the current mean $s(\mathbf{x},B_k)$ have low influence and only affect the weighted error when duplicated many times, which in turn causes $B_k$ to be dominated by repeated copies of such prompts and harms diversity. To encourage diversity in the prompt banks, we penalize larger duplication by normalizing the error change by the duplication factor and using $\Delta\varepsilon(k,t,d)/d$ as the objective in \Cref{eq:dup-opt} and \Cref{eq:best-prompt}.

\section{SAMME.R Algorithm}
\begin{algorithm}[t!]
    \caption{SAMME.R~\cite{zhu2005multi}}
    \label{alg:samme_r}
    \begin{algorithmic}[1]
        \Require
        Training data $\mathcal{D}=\{(\mathbf{x}_i,y_i)\}_{i=1}^n$ with $y_i\in\{1,\dots,K\}$; number of boosting rounds $M$.
        \State (a) \textbf{Initialize} sample weights $w_i \leftarrow 1/n$ for $i=1,\dots,n$, and \textbf{recode} each label $y_i$ into a $K$-dimensional vector $\tilde{\mathbf{y}}_i=(\tilde y_{i,1}, \dots, \tilde y_{i,K})^\top$, where
        \[
        \tilde y_{i,k} =
        \begin{cases}
            1, & \text{if } y_i = k,\\[2pt]
            -\dfrac{1}{K-1}, & \text{if } y_i \neq k.
        \end{cases}
        \]
        \For{$m=1,2,\dots,M$}
            \State (b) \textbf{Fit} a classifier $F^{(m)}$ to the training data $\mathcal D$ using 
            \Statex \qquad the current weights $w_i$, outputting the class 
            \Statex \qquad probability estimates:
            \Statex \[
            F^{(m)}(\mathbf x) = \Big(F^{(m)}_1(\mathbf x),\dots,F^{(m)}_K(\mathbf x)\Big)^\top\;.
            \]
            \State (c) \textbf{Update} sample weights for the next round:
            \Statex \qquad \[
            w_i' = w_i \exp\!\Big(-\frac{K-1}{K}\; \tilde{\mathbf{y}}_i^\top \log F^{(m)}(\mathbf{x}_i)\Big),
            \]\[
            w_i \leftarrow \frac{w_i'}{\sum_{j=1}^{n} w_j'},\quad i=1,\dots,n.
            \]
            \Statex \qquad (here, $\log$ denotes the natural logarithm applied 
            \Statex \qquad element-wise.)
        \EndFor
        \State (d) \textbf{Output} the final strong classifier $S$:
        \[
        S(\mathbf{x}) = \arg\max_{k\in\{1,\dots,K\}}\;\sum_{m=1}^{M} s_k^{(m)}(\mathbf{x})\;,
        \]
        \Statex where, for each round $m$ and class $k=1,\dots,K$, the class scores are defined by
        \[
        s_k^{(m)}(\mathbf{x}) = (K-1)\!\Big( \log F_k^{(m)}(\mathbf{x}) - \frac{1}{K}\sum_{j=1}^{K}\log F_{j}^{(m)}(\mathbf{x})\Big)\;.
        \]
    \end{algorithmic}
\end{algorithm}
SAMME.R~\cite{zhu2005multi} is a multi-class extension of AdaBoost~\cite{freund1997adaboost} that uses real-valued, confidence-rated weak classifiers. It can be derived via forward stagewise additive modeling with a multi-class exponential loss under the symmetric $K$-class label coding. The full procedure is given in \Cref{alg:samme_r}. At initialization (a), sample weights are uniformly set to $1/n$ and each label $y_i$ is recoded into a $K$-dimensional vector $\tilde{\mathbf y}_i$. At boosting round $m$, (b) a classifier is fit to the training data using the current weights, which outputs $K$-class probability estimates under the weighted empirical distribution.
The fitting algorithm is typically a decision tree; in our framework we propose the GPC algorithm. (c) Then the sample weights are updated using the log-probabilities and this fit–reweight cycle repeats. In practice, to ensure that the logarithm is well defined, each probability is clipped to $[\epsilon, 1]$ for a small $\epsilon>0$. After $M$ rounds, (d) the final prediction is obtained by summing the class scores from all rounds and taking the class with the largest total score.

\section{Implementation and Experimental Details}
\label{sup:sec:imp_exp_detail}
\subsection{Hardware Setup and Model Specification}
\subsubsection{Hardware setup.} All experiments run on a single NVIDIA RTX 3090 (24\,GB). For configurations whose GPU memory usage exceeded 24\,GB, we used an RTX A6000 (48\,GB). All other settings were identical across GPUs.
\subsubsection{Model specification.}
We evaluated publicly released CLIP‑style vision-langua\-ge encoders using each implementation’s default tokenizer and preprocessing. The families and variants were: OpenCLIP~\cite{cherti2023openclip} (ViT‑B/16, ViT‑L/14, ViT‑H/14, ViT‑g/14; trained on LAION‑2B~\cite{schuhmann2022laion}); DFN‑CLIP~\cite{fang2023dfn} (ViT‑B/16, ViT‑L/14 on DFN‑2B; ViT‑H/14 on DFN‑5B); CLIPA‑v2~\cite{li2023clipa} (ViT‑L/14, ViT‑H/14; trained on DataComp‑1B~\cite{gadre2023datacomp}); SigLIP2~\cite{tschannen2025siglip2} (ViT‑B/16, ViT‑L/14, ViT‑g/16; trained on WebLI~\cite{chen2022webli}); MetaCLIP2~\cite{chuang2025metaclip2} (ViT‑H/14, ViT‑G/14; trained on curated worldwide public‑web data spanning 300+ languages); and EVA‑02‑CLIP~\cite{sun2023evaclip} (ViT‑B/16 and ViT‑L/14 trained on the Merged‑2B, a mixture of LAION‑2B and COYO‑700\-M~\cite{byeon2022coyo}; ViT‑E/14 trained on LAION‑2B).

\subsection{Baseline Implementation Details}
\subsubsection{PEZ baseline.} PEZ~\cite{wen2023hard} is a gradient-based hard-prompt optimization method that learns discrete text prompts by projecting continuous prompt embeddings onto the nearest vocabulary tokens at each optimization step. For the baseline of gradient-optimized text prompts, we re-implement PEZ for our image-classification setting, following the official implementation. Concretely, we optimize 16 learnable context tokens that are concatenated with the class name, while keeping the vision-language backbone frozen. We train these prompts using the AdamW optimizer~\cite{loshchilov2017adamw} with cross-entropy loss for 200 epochs, using a batch size of 32 and a learning rate of 0.1, closely following the hyperparameter choices reported in the original work.

\subsubsection{EFT baseline.}
We include emulated fine-tuning (EFT)~\cite{mitchell2023eft} as a baseline for transferring adaptation knowledge across models. EFT uses a pre-trained model and its fine-tuned counterpart as source models, and adds the difference between their logits to the target model’s logits so that the target approximately behaves as if it were fine-tuned in the same way. Following the EFT formulation for VLMs in~\cite{park2025transferable}, we adapt EFT to our prompt-learning baselines: given logits $z_{\mathrm{pt\mbox{-}s}}$ and $z_{\mathrm{ft\mbox{-}s}}$ from a pre-trained and prompt-tuned source VLM and logits $z_{\mathrm{pt\mbox{-}t}}$ from a pre-trained target VLM, we form EFT logits $z_{\mathrm{EFT}} = z_{\mathrm{pt\mbox{-}t}} + (z_{\mathrm{ft\mbox{-}s}} - z_{\mathrm{pt\mbox{-}s}})$ for classification. Because this construction requires running three VLMs at inference time (source pre-trained, source prompt-tuned, and target pre-trained), the EFT baseline incurs additional inference cost compared to other text-based transferable baselines.

\section{Additional Experiments and Analyses}
\label{sup:sec:additonal}
\begin{figure}[t]
    \centering
    \includegraphics[width=\linewidth]{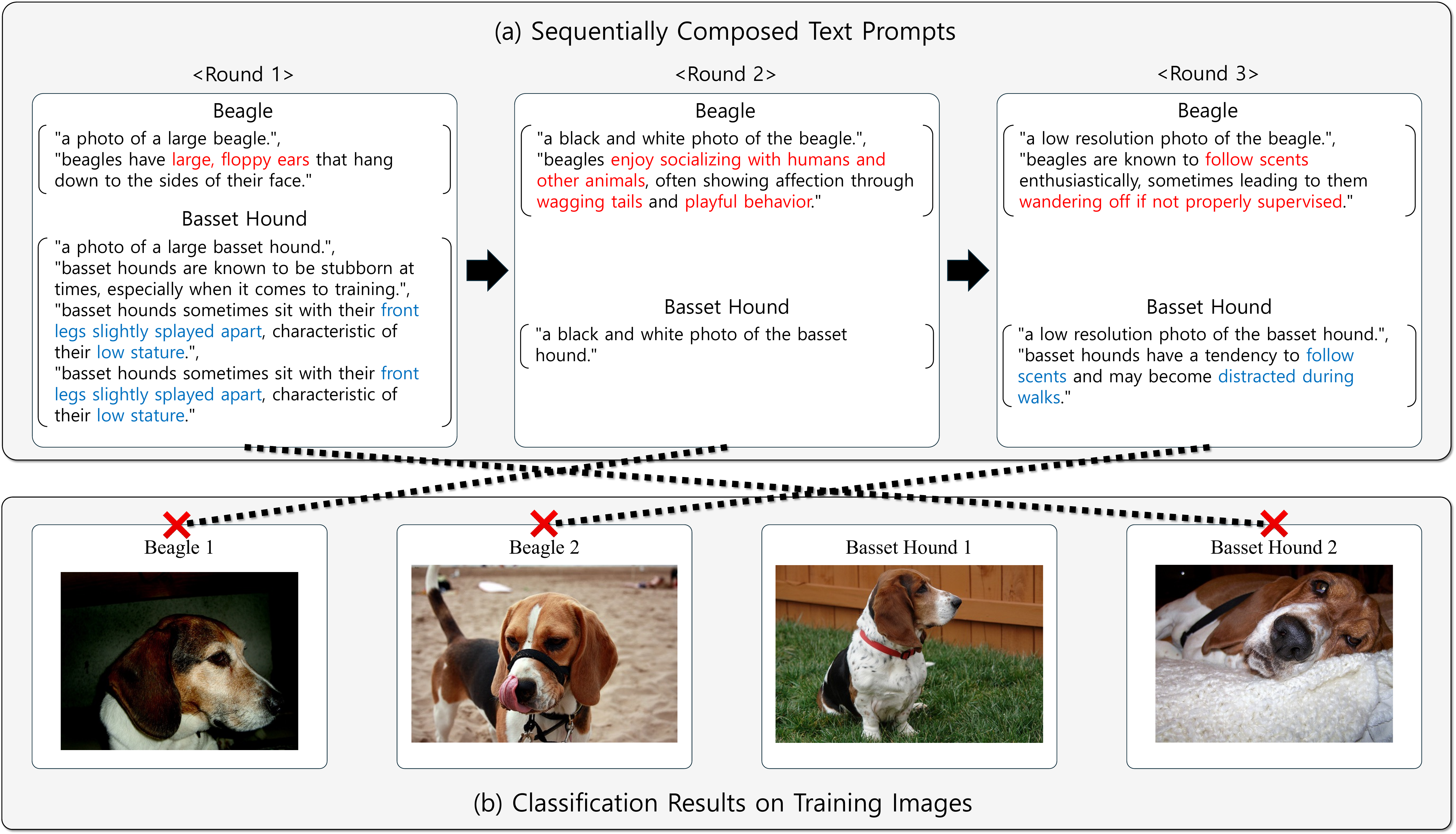}
    \caption{\textbf{Text prompts from consecutive boosting rounds on a two-class toy task.} (a) Sequentially composed prompts for beagle and basset hound over three boosting rounds. (b) Classification results of each classifier on the training images; misclassified images are shown as red crosses along the dotted line for each classifier.}
    \label{fig:appendix_examples}
\end{figure}

\subsection{Text Prompts from Consecutive Boosting Rounds}
\Cref{fig:appendix_examples} presents the prompts selected by GPC across boosting rounds on a two‑class toy task (beagle vs. basset hound). In Round 1, the weak classifier mostly relies on appearance cues, e.g., “large, floppy ears” for beagle and “low stature” with “front legs slightly splayed apart” for basset hound. This round correctly handles most images but fails on Basset Hound 2, where ears and legs are largely occluded by the lying pose. 
In Round 2, the newly composed prompts introduce behavioral cues for beagle; “enjoy socializing with humans and other animals, \dots wagging tails and playful behavior”. This change helps correct the earlier occlusion case; however, Beagle 1 is now misclassified, likely because the image is dark and the dog does not visibly display the “playful” traits suggested by the text.
In Round 3, the prompts add further behavioral cues centered on scent‑following. Both classes emphasize similar aspects: the beagle prompt stresses “follow scents enthusiastically \dots wander off if not properly supervised,” and the basset hound prompt mentions “a tendency to follow scents” and being “distracted during walks.” These cues fix the previous mistakes, yet Beagle 2 is newly misclassified.
We cannot attribute this to a single phrase with high confidence; a plausible factor is that the muzzle and leash dampen visible “sniffing/exploratory” cues, leaving the image weakly aligned with either class description and hence easy to confuse. After three additional boosting rounds, the ensemble correctly classifies all examples in \Cref{fig:appendix_examples}b.

\subsection{Effect of duplication.}
Introducing a learnable weight for each prompt yields too many parameters relative to the few-shot training signal, leading to overfitting and degraded transferability.
To avoid this, GPC uses within-round duplication as a parameter-free approximation of continuous weighting.
As shown in Table~\ref{tab:dup}, even without duplication, TPB still substantially outperforms the strongest baseline, confirming that the core gains come from the boosting framework itself.
Duplication serves as a refinement that further improves performance, with $53\%$ of selected prompts chosen more than once.
\begin{table}[h]
\caption{\textbf{Effect of prompt duplication.} Top-1 accuracy (\%) in the 16-shot setting, averaged over eleven datasets and three seeds.}

\label{tab:dup}
\centering
\resizebox{0.7\linewidth}{!}{
\setlength{\tabcolsep}{4pt}
    \begin{tabular}{cc|c|c|cc}
    \toprule
    \multicolumn{2}{c|}{\multirow{2}{*}{Backbones}} & \multirow{2}{*}{ZS-CLIP} & \multirow{2}{*}{ProAPO} & \multicolumn{2}{c}{\textbf{TPB (Ours)}} \\
    & & & & w/o dup & w/ dup \\
    \midrule
    Source & ViT-B/32 & 60.28 & \acc{67.35}{0.37} & \acc{72.77}{0.32} & \acc{\mathbf{72.87}}{0.44}\\
    \midrule
    \multirow{2}{*}{Target} & ViT-L & 76.76 & \acc{80.01}{0.42} & \acc{81.64}{0.11} & \acc{\mathbf{82.07}}{0.37} \\
    & ViT-H & 78.75 & \acc{82.39}{0.44} & \acc{83.99}{0.16} & \acc{\mathbf{84.24}}{0.49} \\
    \bottomrule
    \end{tabular}
}
\end{table}

\subsection{Prompt pool robustness.}
We restrict the prompt pool to CuPL+DCLIP and evaluate transfer to ViT-L/H models.
Even with this weaker pool, TPB outperforms the baselines, as shown in Table~\ref{tab:pool}, indicating that TPB remains competitive even with a restricted pool, while the full pool provides additional gains.

\begin{table}[h]
\caption{\textbf{Prompt pool ablation.} Top-1 accuracy (\%) in the 16-shot setting, averaged over eleven datasets and three seeds.}
\label{tab:pool}
\centering
\resizebox{0.75\linewidth}{!}{
\setlength{\tabcolsep}{4pt}
\begin{tabular}{c|c|cc|cc}
\toprule
\multirow{2}{*}{Target} & \multirow{2}{*}{ZS-CLIP} & \multirow{2}{*}{PEZ} & \multirow{2}{*}{ProAPO} & \multicolumn{2}{c}{\textbf{TPB (Ours)}} \\
& & & & CuPL+DCLIP & All \\
\midrule
ViT-L       & $76.76$ & \acc{58.93}{2.03} & \acc{80.01}{0.42} & \acc{\underline{81.12}}{0.17} & \acc{\mathbf{82.07}}{0.37} \\
ViT-H       & $78.75$ & \acc{57.60}{1.84} & \acc{82.39}{0.44} & \acc{\underline{82.88}}{0.11} & \acc{\mathbf{84.24}}{0.49} \\
\bottomrule
\end{tabular}
}
\end{table}

\subsection{Large-to-small transfer}
We test reverse-direction transfer from OpenAI ViT-L/14 to EVA-02 B/16 in 16-shot setting on 11 datasets.
TPB yields a $+10.17$\,pp gain on the source model ($71.07\!\rightarrow$\acc{81.24}{0.37}) and retains a $+6.98$\,pp gain after transfer ($69.93\!\rightarrow$\acc{76.91}{0.46}).
This provides initial evidence for TPB's bidirectional transferability.

\subsection{Shot-wise Transfer Performance on Larger Target VLMs}
\Cref{tab:sup_transfer} complements \Cref{tab:transfer} by presenting shot-wise transfer accuracy for larger target VLMs (\eg, ViT-E, ViT-G) when text-based baselines and our method are adapted on OpenAI CLIP ViT-B/32.
As observed in the results of \Cref{tab:transfer}, TPB consistently surpasses the baseline, even when applied to larger VLMs.

\begin{table}[t]
    \caption{
        \textbf{Cross-model transferability of TPB.}
        Average Top-1 accuracy ($\%$) across eleven datasets. All methods are optimized on OpenAI CLIP ViT-B/32 and directly evaluated on much larger, heterogeneous target models (ViT-G and ViT-E). TPB preserves shot-driven gains where other text-based baselines saturate, demonstrating superior robustness and scalability during model transfer.
    }
    \label{tab:sup_transfer}

    \centering
    \footnotesize
    \setlength{\tabcolsep}{4pt}
    
    \scalebox{0.85}{
    \begin{tabular}{ll|ccccc}
        \toprule
        \multirow{2}{*}{\textbf{Target Model}} & \multirow{2}{*}{\textbf{Method}} & \multicolumn{5}{c}{\textbf{Shot}} \\
        & & 1 & 2 & 4 & 8 & 16\\
        
        \midrule
        
        \rowcolor{gray!15}
        \multicolumn{7}{c}{\textit{Source Model: OpenAI CLIP, ViT-B/32}}\\
        
        \midrule
        & ZS-CLIP & \multicolumn{5}{c}{77.18} \\
        \cmidrule(){2-7}
        & PEZ & \acc{50.22}{2.85} & \acc{54.05}{1.85} & \acc{57.95}{1.09} & \acc{59.35}{1.38} & \acc{61.07}{2.01} \\
        & ProAPO & \acc{\underline{79.93}}{0.54} & \acc{\underline{79.90}}{0.51} & \acc{\underline{80.41}}{0.42} & \acc{\underline{80.48}}{0.54} & \acc{\underline{80.45}}{0.36} \\
        \rowcolor{blue!8}
        \cellcolor{white}
        \multirow{-4}{*}{\shortstack[l]{ViT-G: \\ \scriptsize{OpenCLIP}}}
        & \textbf{TPB (Ours)} & \acc{\mathbf{80.18}}{0.41} & \acc{\mathbf{80.94}}{0.95} & \acc{\mathbf{81.44}}{0.61} & \acc{\mathbf{82.63}}{0.54} & \acc{\mathbf{82.74}}{0.41} \\   

        \midrule
        & ZS-CLIP & \multicolumn{5}{c}{83.57} \\
        \cmidrule(){2-7}
        & PEZ & \acc{50.74}{2.07} & \acc{55.31}{2.15} & \acc{57.47}{2.68} & \acc{58.64}{2.52} & \acc{59.54}{1.98} \\
        & ProAPO & \acc{\underline{85.24}}{0.68} & \acc{\underline{85.00}}{0.69} & \acc{\underline{85.31}}{0.38} & \acc{\underline{85.23}}{0.34} & \acc{\underline{85.36}}{0.34} \\
        \rowcolor{blue!8}
        \cellcolor{white}
        \multirow{-4}{*}{\shortstack[l]{ViT-G: \\ \scriptsize{SigLIP2}}}
        & \textbf{TPB (Ours)} & \acc{\mathbf{85.65}}{0.49} & \acc{\mathbf{86.00}}{0.65} & \acc{\mathbf{86.67}}{0.35} & \acc{\mathbf{87.17}}{0.49} & \acc{\mathbf{87.72}}{0.43} \\

        \midrule
        & ZS-CLIP & \multicolumn{5}{c}{81.29} \\
        \cmidrule(){2-7}
        & PEZ & \acc{41.28}{2.44} & \acc{45.25}{2.20} & \acc{47.66}{2.74} & \acc{50.06}{1.59} & \acc{49.71}{2.03} \\
        & ProAPO & \acc{\mathbf{83.51}}{0.93} & \acc{\mathbf{83.19}}{0.86} & \acc{\underline{82.96}}{0.77} & \acc{\underline{83.50}}{0.62} & \acc{\underline{83.27}}{0.52} \\
        \rowcolor{blue!8}
        \cellcolor{white}
        \multirow{-4}{*}{\shortstack[l]{ViT-G: \\ \scriptsize{MetaCLIP2}}}
        & \textbf{TPB (Ours)} & \acc{\underline{83.25}}{0.37} & \acc{\underline{83.10}}{0.95} & \acc{\mathbf{84.02}}{0.38} & \acc{\mathbf{84.45}}{0.60} & \acc{\mathbf{85.00}}{0.60} \\

        \midrule
        & ZS-CLIP & \multicolumn{5}{c}{80.28} \\
        \cmidrule(){2-7}
        & PEZ & \acc{42.53}{2.72} & \acc{46.45}{2.09} & \acc{49.15}{2.23} & \acc{52.66}{1.32} & \acc{52.50}{2.31} \\
        & ProAPO & \acc{\underline{82.81}}{0.58} & \acc{\underline{82.50}}{0.46} & \acc{\underline{83.06}}{0.40} & \acc{\underline{82.95}}{0.38} & \acc{\underline{83.18}}{0.46} \\
        \rowcolor{blue!8}
        \cellcolor{white}
        \multirow{-4}{*}{\shortstack[l]{ViT-G: \\ \scriptsize{CLIPA-v2}}}
        & \textbf{TPB (Ours)} & \acc{\mathbf{82.85}}{0.37} & \acc{\mathbf{82.87}}{0.80} & \acc{\mathbf{83.89}}{0.30} & \acc{\mathbf{84.34}}{0.39} & \acc{\mathbf{84.94}}{0.48} \\

        \midrule
        & ZS-CLIP & \multicolumn{5}{c}{79.53} \\
        \cmidrule(){2-7}
        & PEZ & \acc{50.73}{1.63} & \acc{55.93}{1.91} & \acc{58.79}{1.95} & \acc{60.64}{1.24} & \acc{62.67}{2.09} \\
        & ProAPO & \acc{\underline{82.65}}{0.54} & \acc{\underline{82.70}}{0.52} & \acc{\underline{83.01}}{0.67} & \acc{\underline{83.11}}{0.50} & \acc{\underline{83.10}}{0.42} \\
        \rowcolor{blue!8}
        \cellcolor{white}
        \multirow{-4}{*}{\shortstack[l]{ViT-E: \\ \scriptsize{EVA-02-CLIP}}}
        & \textbf{TPB (Ours)} & \acc{\mathbf{82.85}}{0.44} & \acc{\mathbf{83.18}}{0.52} & \acc{\mathbf{84.26}}{0.35} & \acc{\mathbf{84.60}}{0.40} & \acc{\mathbf{85.37}}{0.43} \\
        \bottomrule
    \end{tabular}
    }
\end{table}

\section{Full Experiments Results}
\label{sup:sec:full_results}

\subsection{Full shot-wise scalability.}
{
    \providecommand{\myangle}{90}
    \providecommand{\gr}{\color{gray!70}}
    \begin{table}[h!]
        \caption{
            \textbf{Shot scalability on OpenAI CLIP RN50.} Top-1 accuracy (\%) over eleven datasets. For few-shot methods with seed-level runs, entries are reported as \acc{mean}{std}, where std is computed over seeds; zero-shot methods and LLMbo are shown without std when seed-level runs are unavailable.
        }
        \label{tab:mainshot_full_rn50}
        
        \centering
        \footnotesize
        \setlength{\tabcolsep}{2pt}

        \scalebox{0.55}{
        \begin{tabular}{cl|ccccccccccc|c}
            \toprule
            \textbf{Shot} & \textbf{Method} & \rotatebox{\myangle}{\textbf{IN-1K}} & \rotatebox{\myangle}{\textbf{Caltech}} & \rotatebox{\myangle}{\textbf{Pets}} & \rotatebox{\myangle}{\textbf{Cars}} & \rotatebox{\myangle}{\textbf{Flowers}} & \rotatebox{\myangle}{\textbf{Food}} & \rotatebox{\myangle}{\textbf{Aircraft}} & \rotatebox{\myangle}{\textbf{SUN}} & \rotatebox{\myangle}{\textbf{DTD}} & \rotatebox{\myangle}{\textbf{ESAT}} & \rotatebox{\myangle}{\textbf{UCF}} & \rotatebox{\myangle}{\textbf{Avg.}} \\
            
            \midrule
            \rowcolor{gray!15}
            \multicolumn{14}{c}{\textit{Zero-shot scenario}}\\
            \midrule
            
            \multirow{3}{*}{\rotatebox{90}{ZS}} & CLIP & 58.3 & 85.8 & 83.7 & 55.9 & 61.2 & 75.2 & 14.5 & 58.5 & 40.0 & 24.2 & 58.4 & 56.0 \\
             & DCLIP & 59.6 & 88.6 & 83.1 & 53.9 & 66.3 & 76.6 & 16.8 & 61.0 & 41.7 & 37.6 & 60.8 & 58.7 \\
             & CuPL & 61.5 & 88.0 & 87.6 & 56.1 & 68.1 & 77.1 & 18.3 & 61.9 & 47.6 & 36.3 & 62.1 & 60.4 \\
            
            \midrule
            \rowcolor{gray!15}
            \multicolumn{14}{c}{\textit{Few-shot scenario}}\\
            \midrule
            
            \multirow{5}{*}{\rotatebox{90}{1 shot}} & CoOp & \acc{55.5}{1.29} & \acc{88.0}{0.62} & \acc{86.3}{0.68} & \acc{55.6}{1.57} & \acc{68.2}{0.53} & \acc{74.2}{1.28} & \acc{8.4}{5.63} & \acc{60.2}{0.08} & \acc{43.3}{2.71} & \acc{51.1}{5.44} & \acc{61.9}{0.58} & \acc{59.3}{1.86} \\
            \cmidrule{2-14}
             & PEZ & \acc{34.2}{0.46} & \acc{66.5}{1.04} & \acc{68.8}{4.56} & \acc{38.2}{0.50} & \acc{57.8}{1.55} & \acc{51.9}{1.77} & \acc{12.8}{0.43} & \acc{35.6}{0.52} & \acc{28.8}{0.84} & \acc{29.4}{6.65} & \acc{41.8}{1.08} & \acc{42.4}{1.76} \\
             & LLMbo & 59.6 & $\mathbf{89.1}$ & \underline{88.1} & 56.2 & 67.2 & $\mathbf{78.3}$ & 18.1 & 61.0 & 44.8 & 49.0 & 60.2 & 61.1 \\
             & ProAPO & \acc{\mathbf{61.1}}{0.21} & \acc{\underline{89.0}}{0.28} & \acc{\mathbf{89.0}}{0.25} & \acc{\mathbf{57.6}}{0.33} & \acc{\underline{68.9}}{0.41} & \acc{\underline{78.3}}{0.39} & \acc{\underline{18.2}}{0.11} & \acc{\underline{61.9}}{0.18} & \acc{\underline{48.3}}{1.68} & \acc{\mathbf{53.3}}{1.28} & \acc{\underline{63.6}}{0.27} & \acc{\underline{62.7}}{0.49} \\
            \rowcolor{blue!8}
            \cellcolor{white} & \textbf{TPB (Ours)} & \acc{\underline{60.7}}{0.19} & \acc{88.8}{0.56} & \acc{87.2}{0.45} & \acc{\underline{57.4}}{0.47} & \acc{\mathbf{77.0}}{0.32} & \acc{76.4}{0.18} & \acc{\mathbf{19.7}}{0.22} & \acc{\mathbf{63.7}}{0.21} & \acc{\mathbf{53.1}}{2.31} & \acc{\underline{51.5}}{3.33} & \acc{\mathbf{66.2}}{1.04} & \acc{\mathbf{63.8}}{0.84} \\
            
            \midrule
            
            \multirow{5}{*}{\rotatebox{90}{16 shot}} & CoOp & \acc{60.2}{0.26} & \acc{91.9}{0.29} & \acc{86.3}{0.31} & \acc{72.9}{0.18} & \acc{94.7}{0.17} & \acc{74.4}{0.08} & \acc{31.5}{0.63} & \acc{68.5}{0.19} & \acc{63.4}{0.50} & \acc{83.0}{1.75} & \acc{75.7}{0.28} & \acc{73.0}{0.42} \\
            \cmidrule{2-14}
             & PEZ & \acc{54.5}{0.16} & \acc{85.6}{1.86} & \acc{82.3}{0.21} & \acc{56.9}{0.46} & \acc{\underline{75.5}}{0.51} & \acc{70.5}{0.88} & \acc{\underline{20.6}}{0.31} & \acc{57.0}{0.31} & \acc{50.6}{0.77} & \acc{58.1}{2.87} & \acc{60.4}{0.36} & \acc{61.1}{0.79} \\
             & LLMbo & 59.9 & 89.5 & 88.3 & 56.8 & 67.4 & 78.3 & 18.1 & 60.8 & 44.9 & 51.4 & 60.5 & 61.4 \\
             & ProAPO & \acc{\underline{60.1}}{0.00} & \acc{\underline{90.1}}{0.24} & \acc{\mathbf{89.1}}{0.11} & \acc{\underline{58.5}}{0.22} & \acc{73.3}{0.81} & \acc{\underline{78.8}}{0.08} & \acc{18.7}{0.29} & \acc{\underline{63.0}}{0.15} & \acc{\underline{52.5}}{1.01} & \acc{\underline{58.6}}{2.99} & \acc{\underline{65.5}}{0.61} & \acc{\underline{64.4}}{0.59} \\
            \rowcolor{blue!8}
            \cellcolor{white} & \textbf{TPB (Ours)} & \acc{\mathbf{64.7}}{0.13} & \acc{\mathbf{91.8}}{0.18} & \acc{\underline{89.0}}{0.57} & \acc{\mathbf{65.0}}{0.24} & \acc{\mathbf{85.9}}{0.52} & \acc{\mathbf{79.0}}{0.20} & \acc{\mathbf{24.8}}{0.39} & \acc{\mathbf{70.7}}{0.06} & \acc{\mathbf{62.8}}{0.35} & \acc{\mathbf{66.9}}{2.58} & \acc{\mathbf{73.0}}{0.40} & \acc{\mathbf{70.3}}{0.51} \\
            \bottomrule
        \end{tabular}
        }
    \end{table}
}

{
    \providecommand{\myangle}{90}
    \providecommand{\gr}{\color{gray!70}}
    \begin{table}[h!]
        \caption{
            \textbf{Shot scalability on OpenAI CLIP ViT-B/32.} Top-1 accuracy (\%) over eleven datasets. For few-shot methods with seed-level runs, entries are reported as \acc{mean}{std}, where std is computed over seeds; zero-shot methods are deterministic and shown without std.
        }
        \label{tab:mainshot_full_vitb32}
        
        \centering
        \footnotesize
        \setlength{\tabcolsep}{2pt}

        \scalebox{0.55}{
        \begin{tabular}{cl|ccccccccccc|c}
            \toprule
            \textbf{Shot} & \textbf{Method} & \rotatebox{\myangle}{\textbf{IN-1K}} & \rotatebox{\myangle}{\textbf{Caltech}} & \rotatebox{\myangle}{\textbf{Pets}} & \rotatebox{\myangle}{\textbf{Cars}} & \rotatebox{\myangle}{\textbf{Flowers}} & \rotatebox{\myangle}{\textbf{Food}} & \rotatebox{\myangle}{\textbf{Aircraft}} & \rotatebox{\myangle}{\textbf{SUN}} & \rotatebox{\myangle}{\textbf{DTD}} & \rotatebox{\myangle}{\textbf{ESAT}} & \rotatebox{\myangle}{\textbf{UCF}} & \rotatebox{\myangle}{\textbf{Avg.}} \\
            
            \midrule
            \rowcolor{gray!15}
            \multicolumn{14}{c}{\textit{Zero-shot scenario}}\\
            \midrule
            
            \multirow{3}{*}{\rotatebox{90}{ZS}} & CLIP & 62.1 & 91.2 & 85.0 & 60.4 & 63.8 & 79.2 & 17.8 & 62.0 & 42.8 & 38.2 & 60.7 & 60.3 \\
             & DCLIP & 63.2 & 92.3 & 83.6 & 59.0 & 66.7 & 80.3 & 19.7 & 64.4 & 43.9 & 48.9 & 64.4 & 62.4 \\
             & CuPL & 64.7 & 92.5 & 89.1 & 60.7 & 69.8 & 80.4 & 19.2 & 64.9 & 49.5 & 47.6 & 66.0 & 64.0 \\
            
            \midrule
            \rowcolor{gray!15}
            \multicolumn{14}{c}{\textit{Few-shot scenario}}\\
            \midrule
            
            \multirow{4}{*}{\rotatebox{90}{1 shot}} & CoOp & \acc{59.4}{1.51} & \acc{91.7}{0.88} & \acc{87.6}{1.04} & \acc{59.5}{0.77} & \acc{71.3}{1.60} & \acc{75.5}{2.05} & \acc{15.6}{7.34} & \acc{63.7}{1.11} & \acc{46.3}{1.33} & \acc{55.0}{2.47} & \acc{66.2}{1.01} & \acc{62.9}{1.92} \\
            \cmidrule{2-14}
             & PEZ & \acc{37.6}{0.58} & \acc{74.6}{2.76} & \acc{68.4}{2.66} & \acc{41.1}{0.66} & \acc{62.4}{1.43} & \acc{56.4}{0.14} & \acc{15.3}{0.81} & \acc{40.0}{1.28} & \acc{31.2}{1.02} & \acc{33.2}{4.83} & \acc{48.8}{2.74} & \acc{46.3}{1.72} \\
             & ProAPO & \acc{\mathbf{64.1}}{0.06} & \acc{\mathbf{93.4}}{0.50} & \acc{\mathbf{89.8}}{0.51} & \acc{\mathbf{61.3}}{0.07} & \acc{\underline{72.7}}{0.32} & \acc{\mathbf{81.7}}{0.10} & \acc{\underline{21.5}}{0.87} & \acc{\underline{66.4}}{0.13} & \acc{\mathbf{52.8}}{0.76} & \acc{\underline{56.9}}{1.18} & \acc{\underline{66.5}}{0.13} & \acc{\underline{66.1}}{0.42} \\
            \rowcolor{blue!8}
            \cellcolor{white} & \textbf{TPB (Ours)} & \acc{\underline{63.9}}{0.23} & \acc{\underline{93.0}}{0.75} & \acc{\underline{89.0}}{0.14} & \acc{\underline{60.5}}{0.55} & \acc{\mathbf{79.6}}{0.55} & \acc{\underline{79.6}}{0.30} & \acc{\mathbf{22.5}}{0.95} & \acc{\mathbf{67.4}}{0.21} & \acc{\underline{52.5}}{2.00} & \acc{\mathbf{58.9}}{3.26} & \acc{\mathbf{71.0}}{0.50} & \acc{\mathbf{67.1}}{0.86} \\
            
            \midrule
            
            \multirow{4}{*}{\rotatebox{90}{16 shot}} & CoOp & \acc{66.8}{0.18} & \acc{95.0}{0.17} & \acc{88.5}{0.31} & \acc{76.1}{0.40} & \acc{95.2}{0.42} & \acc{78.5}{0.23} & \acc{33.5}{0.38} & \acc{72.5}{0.20} & \acc{65.4}{0.49} & \acc{83.3}{0.24} & \acc{79.2}{0.49} & \acc{75.8}{0.32} \\
            \cmidrule{2-14}
             & PEZ & \acc{57.4}{0.16} & \acc{90.4}{0.66} & \acc{83.1}{0.97} & \acc{60.2}{0.37} & \acc{\underline{78.0}}{0.61} & \acc{74.0}{0.26} & \acc{21.4}{0.45} & \acc{61.2}{0.41} & \acc{\underline{54.3}}{0.30} & \acc{59.8}{1.09} & \acc{66.4}{1.20} & \acc{64.2}{0.59} \\
             & ProAPO & \acc{\underline{63.5}}{0.20} & \acc{\underline{93.7}}{0.10} & \acc{\underline{90.5}}{0.19} & \acc{\underline{61.9}}{0.16} & \acc{77.7}{0.42} & \acc{\underline{81.8}}{0.05} & \acc{\underline{22.1}}{0.59} & \acc{\underline{66.8}}{0.19} & \acc{53.4}{0.58} & \acc{\underline{60.5}}{0.87} & \acc{\underline{69.1}}{0.72} & \acc{\underline{67.4}}{0.37} \\
            \rowcolor{blue!8}
            \cellcolor{white} & \textbf{TPB (Ours)} & \acc{\mathbf{68.0}}{0.21} & \acc{\mathbf{94.9}}{0.23} & \acc{\mathbf{90.6}}{0.43} & \acc{\mathbf{67.5}}{0.21} & \acc{\mathbf{87.0}}{0.30} & \acc{\mathbf{81.9}}{0.29} & \acc{\mathbf{26.9}}{0.53} & \acc{\mathbf{73.4}}{0.13} & \acc{\mathbf{63.9}}{0.65} & \acc{\mathbf{70.8}}{1.82} & \acc{\mathbf{76.8}}{0.08} & \acc{\mathbf{72.9}}{0.44} \\
            \bottomrule
        \end{tabular}
        }
    \end{table}
}

\Cref{tab:mainshot_full_rn50} and \Cref{tab:mainshot_full_vitb32} reports the complete per-dataset shot-scalability results on the source models.
The full results confirm the trend observed in the main paper: TPB achieves strong average performance among text-based methods and benefits more clearly from additional shots than prior text-prompting baselines.
The improvements are observed across diverse recognition domains, indicating that the shot-scalability of TPB is not driven by a single dataset.

\subsection{Full shot-wise and target-wise transfer.}
\Cref{tab:transfer_by_model_rn50} and \Cref{tab:transfer_by_model_vitb32} expands the grouped transfer results in the main paper by reporting shot-wise performance for each target model. 
This target-wise breakdown confirms that TPB consistently improves over text-prompt baselines such as PEZ and ProAPO, while also revealing a more nuanced comparison with EFT-based prompt-learning baselines.
Although CoOp+EFT and PromptSRC+EFT can be competitive, and occasionally outperform TPB at 16 shots on some target models, their behavior is highly target-dependent; in particular, they suffer severe degradation on DFN targets, suggesting that logit-space emulation can be brittle under changes in architecture or training recipe.
In contrast, TPB transfers by simply re-embedding natural-language prompt ensembles on the target VLM, yielding stable performance across heterogeneous ViT-L and ViT-H targets.
This makes TPB a robust and lightweight transfer option, requiring only the target VLM at inference rather than the multi-model inference used by EFT-based baselines.
{
    \begin{table}[h!]
        \caption{
            \textbf{Target-model-wise transfer from OpenAI CLIP RN50.} Average Top-1 accuracy (\%) across eleven datasets. All few-shot methods are optimized on OpenAI CLIP RN50 and directly evaluated on larger heterogeneous ViT-L and ViT-H target models. Entries are reported as \acc{mean}{std}, where std is the average of per-dataset seed standard deviations; ZS-CLIP is deterministic and shown without std.
        }
        \label{tab:transfer_by_model_rn50}

        \centering
        \scriptsize
        \setlength{\tabcolsep}{3pt}
        \renewcommand{\arraystretch}{0.92}

        \scalebox{0.9}{
        \begin{tabular}{ll|ccccc}
            \toprule
            \multirow{2}{*}{\textbf{Target Model}} & \multirow{2}{*}{\textbf{Method}} & \multicolumn{5}{c}{\textbf{Shot}} \\
            & & 1 & 2 & 4 & 8 & 16 \\
            \midrule
            \rowcolor{gray!15}
            \multicolumn{7}{c}{\textit{Target Models: ViT-L family}}\\
            \midrule
            & ZS-CLIP & \multicolumn{5}{c}{73.35} \\
            \cmidrule(){2-7}
            & CoOp+EFT & \acc{72.51}{1.31} & \acc{73.74}{0.93} & \acc{76.45}{0.83} & \acc{\underline{78.83}}{0.48} & \acc{\mathbf{80.18}}{0.37} \\
            \cmidrule(){2-7}
            & PEZ & \acc{45.88}{2.17} & \acc{48.33}{2.23} & \acc{52.19}{2.03} & \acc{54.86}{2.58} & \acc{56.37}{1.89} \\
            & ProAPO & \acc{\mathbf{76.33}}{0.53} & \acc{\underline{76.86}}{0.56} & \acc{\underline{76.86}}{0.49} & \acc{77.12}{1.02} & \acc{77.93}{0.64} \\
            \rowcolor{blue!8}
            \cellcolor{white}
            \multirow{-5}{*}{\shortstack[l]{ViT-L: \\ OpenCLIP}}
            & \textbf{TPB (Ours)} & \acc{\underline{76.06}}{0.70} & \acc{\mathbf{77.31}}{0.62} & \acc{\mathbf{78.18}}{0.56} & \acc{\mathbf{79.11}}{0.55} & \acc{\underline{79.73}}{0.52} \\
            \midrule
            & ZS-CLIP & \multicolumn{5}{c}{81.44} \\
            \cmidrule(){2-7}
            & CoOp+EFT & \acc{77.87}{2.23} & \acc{78.85}{0.96} & \acc{81.57}{0.82} & \acc{\underline{83.27}}{0.34} & \acc{\underline{84.72}}{0.35} \\
            \cmidrule(){2-7}
            & PEZ & \acc{49.25}{2.01} & \acc{53.38}{1.51} & \acc{56.77}{2.27} & \acc{58.17}{1.53} & \acc{59.28}{1.44} \\
            & ProAPO & \acc{\underline{81.92}}{0.50} & \acc{\underline{82.46}}{0.40} & \acc{\underline{82.21}}{0.48} & \acc{82.68}{0.65} & \acc{82.89}{0.43} \\
            \rowcolor{blue!8}
            \cellcolor{white}
            \multirow{-5}{*}{\shortstack[l]{ViT-L: \\ SigLIP2}}
            & \textbf{TPB (Ours)} & \acc{\mathbf{83.32}}{0.60} & \acc{\mathbf{83.92}}{0.41} & \acc{\mathbf{84.52}}{0.43} & \acc{\mathbf{84.69}}{0.63} & \acc{\mathbf{85.10}}{0.47} \\
            \midrule
            & ZS-CLIP & \multicolumn{5}{c}{77.22} \\
            \cmidrule(){2-7}
            & CoOp+EFT & \acc{22.59}{3.22} & \acc{25.97}{2.87} & \acc{31.43}{3.58} & \acc{35.90}{1.45} & \acc{39.65}{1.58} \\
            \cmidrule(){2-7}
            & PEZ & \acc{53.85}{2.02} & \acc{57.90}{1.90} & \acc{59.11}{2.34} & \acc{62.27}{1.41} & \acc{63.46}{1.74} \\
            & ProAPO & \acc{\underline{78.64}}{0.88} & \acc{\underline{79.76}}{0.51} & \acc{\underline{79.64}}{0.58} & \acc{\underline{80.50}}{0.69} & \acc{\underline{80.20}}{0.50} \\
            \rowcolor{blue!8}
            \cellcolor{white}
            \multirow{-5}{*}{\shortstack[l]{ViT-L: \\ DFN}}
            & \textbf{TPB (Ours)} & \acc{\mathbf{79.84}}{0.75} & \acc{\mathbf{80.73}}{0.41} & \acc{\mathbf{81.51}}{0.34} & \acc{\mathbf{81.85}}{0.39} & \acc{\mathbf{82.19}}{0.50} \\
            \midrule
            & ZS-CLIP & \multicolumn{5}{c}{74.96} \\
            \cmidrule(){2-7}
            & CoOp+EFT & \acc{71.22}{1.80} & \acc{73.96}{1.28} & \acc{77.04}{0.84} & \acc{\underline{79.25}}{0.38} & \acc{\mathbf{81.01}}{0.44} \\
            \cmidrule(){2-7}
            & PEZ & \acc{49.63}{2.48} & \acc{54.57}{1.84} & \acc{58.18}{1.68} & \acc{60.83}{1.59} & \acc{62.68}{1.80} \\
            & ProAPO & \acc{\underline{77.13}}{0.92} & \acc{\underline{77.90}}{0.42} & \acc{\underline{77.59}}{0.80} & \acc{77.78}{1.03} & \acc{78.83}{0.57} \\
            \rowcolor{blue!8}
            \cellcolor{white}
            \multirow{-5}{*}{\shortstack[l]{ViT-L: \\ EVA-02-CLIP}}
            & \textbf{TPB (Ours)} & \acc{\mathbf{77.65}}{0.80} & \acc{\mathbf{78.49}}{0.46} & \acc{\mathbf{79.69}}{0.68} & \acc{\mathbf{80.18}}{0.61} & \acc{\underline{80.82}}{0.45} \\
            \midrule
            & ZS-CLIP & \multicolumn{5}{c}{76.82} \\
            \cmidrule(){2-7}
            & CoOp+EFT & \acc{74.44}{1.69} & \acc{76.41}{1.06} & \acc{78.94}{0.74} & \acc{\underline{80.83}}{0.44} & \acc{\mathbf{82.04}}{0.38} \\
            \cmidrule(){2-7}
            & PEZ & \acc{40.20}{2.28} & \acc{44.59}{2.41} & \acc{47.91}{2.49} & \acc{49.31}{2.22} & \acc{50.70}{2.43} \\
            & ProAPO & \acc{\underline{78.26}}{1.14} & \acc{\underline{78.97}}{0.53} & \acc{\underline{79.17}}{0.89} & \acc{79.21}{1.14} & \acc{79.62}{0.61} \\
            \rowcolor{blue!8}
            \cellcolor{white}
            \multirow{-5}{*}{\shortstack[l]{ViT-L: \\ CLIPA-v2}}
            & \textbf{TPB (Ours)} & \acc{\mathbf{78.70}}{0.45} & \acc{\mathbf{79.68}}{0.37} & \acc{\mathbf{80.40}}{0.64} & \acc{\mathbf{81.02}}{0.48} & \acc{\underline{81.51}}{0.38} \\
            \midrule
            \rowcolor{gray!15}
            \multicolumn{7}{c}{\textit{Target Models: ViT-H family}}\\
            \midrule
            & ZS-CLIP & \multicolumn{5}{c}{76.51} \\
            \cmidrule(){2-7}
            & CoOp+EFT & \acc{75.63}{1.10} & \acc{77.08}{0.82} & \acc{79.03}{0.53} & \acc{\underline{81.03}}{0.31} & \acc{\underline{82.10}}{0.25} \\
            \cmidrule(){2-7}
            & PEZ & \acc{47.56}{2.33} & \acc{51.57}{2.02} & \acc{54.71}{2.21} & \acc{57.82}{2.06} & \acc{59.43}{1.47} \\
            & ProAPO & \acc{\underline{78.47}}{0.79} & \acc{\underline{78.85}}{0.62} & \acc{\underline{79.46}}{0.57} & \acc{79.51}{1.24} & \acc{80.13}{0.92} \\
            \rowcolor{blue!8}
            \cellcolor{white}
            \multirow{-5}{*}{\shortstack[l]{ViT-H: \\ OpenCLIP}}
            & \textbf{TPB (Ours)} & \acc{\mathbf{79.52}}{0.85} & \acc{\mathbf{80.46}}{0.58} & \acc{\mathbf{81.42}}{0.28} & \acc{\mathbf{82.00}}{0.32} & \acc{\mathbf{82.58}}{0.38} \\
            \midrule
            & ZS-CLIP & \multicolumn{5}{c}{81.51} \\
            \cmidrule(){2-7}
            & CoOp+EFT & \acc{23.50}{3.38} & \acc{26.66}{3.06} & \acc{32.21}{3.73} & \acc{36.87}{1.49} & \acc{40.49}{1.57} \\
            \cmidrule(){2-7}
            & PEZ & \acc{56.33}{2.19} & \acc{59.80}{1.97} & \acc{63.63}{2.32} & \acc{65.58}{1.85} & \acc{66.94}{1.63} \\
            & ProAPO & \acc{\mathbf{83.79}}{0.32} & \acc{\underline{84.00}}{0.40} & \acc{\underline{83.92}}{0.53} & \acc{\underline{84.36}}{0.72} & \acc{\underline{84.54}}{0.51} \\
            \rowcolor{blue!8}
            \cellcolor{white}
            \multirow{-5}{*}{\shortstack[l]{ViT-H: \\ DFN}}
            & \textbf{TPB (Ours)} & \acc{\underline{83.34}}{0.56} & \acc{\mathbf{84.09}}{0.47} & \acc{\mathbf{84.81}}{0.33} & \acc{\mathbf{85.57}}{0.39} & \acc{\mathbf{85.46}}{0.40} \\
            \midrule
            & ZS-CLIP & \multicolumn{5}{c}{78.50} \\
            \cmidrule(){2-7}
            & CoOp+EFT & \acc{77.05}{1.72} & \acc{78.45}{0.72} & \acc{80.54}{0.67} & \acc{\underline{82.39}}{0.34} & \acc{\mathbf{83.58}}{0.39} \\
            \cmidrule(){2-7}
            & PEZ & \acc{41.14}{2.42} & \acc{46.02}{2.41} & \acc{48.72}{2.75} & \acc{51.13}{3.22} & \acc{52.19}{2.12} \\
            & ProAPO & \acc{\underline{80.13}}{0.67} & \acc{\underline{80.92}}{0.55} & \acc{\underline{81.09}}{0.97} & \acc{81.33}{0.86} & \acc{81.80}{0.52} \\
            \rowcolor{blue!8}
            \cellcolor{white}
            \multirow{-5}{*}{\shortstack[l]{ViT-H: \\ CLIPA-v2}}
            & \textbf{TPB (Ours)} & \acc{\mathbf{80.77}}{0.80} & \acc{\mathbf{81.61}}{0.54} & \acc{\mathbf{82.34}}{0.49} & \acc{\mathbf{82.92}}{0.34} & \acc{\underline{83.13}}{0.33} \\
            \midrule
            & ZS-CLIP & \multicolumn{5}{c}{78.47} \\
            \cmidrule(){2-7}
            & CoOp+EFT & \acc{72.40}{2.10} & \acc{74.45}{1.75} & \acc{77.99}{1.06} & \acc{80.28}{0.40} & \acc{81.82}{0.37} \\
            \cmidrule(){2-7}
            & PEZ & \acc{39.62}{1.97} & \acc{42.50}{2.20} & \acc{46.46}{1.70} & \acc{48.69}{2.47} & \acc{50.00}{2.34} \\
            & ProAPO & \acc{\mathbf{81.49}}{0.57} & \acc{\underline{81.50}}{0.94} & \acc{\underline{81.68}}{0.51} & \acc{\underline{81.75}}{0.78} & \acc{\underline{82.40}}{0.84} \\
            \rowcolor{blue!8}
            \cellcolor{white}
            \multirow{-5}{*}{\shortstack[l]{ViT-H: \\ MetaCLIP2}}
            & \textbf{TPB (Ours)} & \acc{\underline{81.02}}{0.70} & \acc{\mathbf{82.07}}{0.56} & \acc{\mathbf{82.82}}{0.23} & \acc{\mathbf{83.57}}{0.54} & \acc{\mathbf{83.87}}{0.29} \\
            \bottomrule
        \end{tabular}
        }
    \end{table}
}

{
    \begin{table}[h!]
        \caption{
            \textbf{Target-model-wise transfer from OpenAI CLIP ViT-B/32.} Average Top-1 accuracy (\%) across eleven datasets. All few-shot methods are optimized on OpenAI CLIP ViT-B/32 and directly evaluated on larger heterogeneous ViT-L and ViT-H target models. Entries are reported as \acc{mean}{std}, where std is the average of per-dataset seed standard deviations; ZS-CLIP is deterministic and shown without std.
        }
        \label{tab:transfer_by_model_vitb32}

        \centering
        \scriptsize
        \setlength{\tabcolsep}{3pt}
        \renewcommand{\arraystretch}{0.92}

        \scalebox{0.9}{
        \begin{tabular}{ll|ccccc}
            \toprule
            \multirow{2}{*}{\textbf{Target Model}} & \multirow{2}{*}{\textbf{Method}} & \multicolumn{5}{c}{\textbf{Shot}} \\
            & & 1 & 2 & 4 & 8 & 16 \\
            \midrule
            \rowcolor{gray!15}
            \multicolumn{7}{c}{\textit{Target Models: ViT-L family}}\\
            \midrule
            & ZS-CLIP & \multicolumn{5}{c}{73.35} \\
            \cmidrule(){2-7}
            & CoOp+EFT & \acc{72.86}{1.55} & \acc{74.85}{1.19} & \acc{77.05}{0.76} & \acc{\underline{79.07}}{0.43} & \acc{\mathbf{80.71}}{0.25} \\
            & PromptSRC+EFT & \acc{74.84}{1.04} & \acc{75.73}{0.88} & \acc{\underline{78.10}}{0.53} & \acc{79.02}{0.53} & \acc{80.28}{0.24} \\
            \cmidrule(){2-7}
            & PEZ & \acc{46.82}{2.60} & \acc{50.65}{2.61} & \acc{53.46}{2.19} & \acc{54.95}{1.96} & \acc{57.40}{1.77} \\
            & ProAPO & \acc{\mathbf{77.23}}{0.74} & \acc{\underline{77.37}}{0.56} & \acc{77.51}{0.63} & \acc{77.94}{0.49} & \acc{77.97}{0.44} \\
            \rowcolor{blue!8}
            \cellcolor{white}
            \multirow{-6}{*}{\shortstack[l]{ViT-L: \\ OpenCLIP}}
            & \textbf{TPB (Ours)} & \acc{\underline{77.12}}{0.50} & \acc{\mathbf{77.54}}{1.14} & \acc{\mathbf{78.92}}{0.62} & \acc{\mathbf{79.76}}{0.53} & \acc{\underline{80.30}}{0.35} \\
            \midrule
            & ZS-CLIP & \multicolumn{5}{c}{81.44} \\
            \cmidrule(){2-7}
            & CoOp+EFT & \acc{76.90}{2.31} & \acc{78.51}{1.52} & \acc{81.29}{0.92} & \acc{83.00}{0.52} & \acc{84.74}{0.44} \\
            & PromptSRC+EFT & \acc{79.87}{1.44} & \acc{80.42}{1.34} & \acc{\underline{83.27}}{0.62} & \acc{\underline{84.46}}{0.64} & \acc{\underline{85.16}}{0.43} \\
            \cmidrule(){2-7}
            & PEZ & \acc{49.94}{1.26} & \acc{55.23}{1.35} & \acc{57.33}{1.55} & \acc{58.36}{1.87} & \acc{59.05}{2.69} \\
            & ProAPO & \acc{\underline{82.57}}{0.66} & \acc{\underline{82.50}}{0.66} & \acc{82.64}{0.49} & \acc{82.58}{0.50} & \acc{82.59}{0.52} \\
            \rowcolor{blue!8}
            \cellcolor{white}
            \multirow{-6}{*}{\shortstack[l]{ViT-L: \\ SigLIP2}}
            & \textbf{TPB (Ours)} & \acc{\mathbf{83.59}}{0.33} & \acc{\mathbf{84.03}}{0.66} & \acc{\mathbf{84.49}}{0.41} & \acc{\mathbf{85.34}}{0.32} & \acc{\mathbf{85.46}}{0.36} \\
            \midrule
            & ZS-CLIP & \multicolumn{5}{c}{77.22} \\
            \cmidrule(){2-7}
            & CoOp+EFT & \acc{23.73}{4.70} & \acc{26.89}{4.12} & \acc{33.02}{3.70} & \acc{36.60}{2.22} & \acc{41.19}{1.34} \\
            & PromptSRC+EFT & \acc{27.80}{3.61} & \acc{30.78}{3.31} & \acc{37.76}{4.45} & \acc{41.95}{2.21} & \acc{46.09}{1.72} \\
            \cmidrule(){2-7}
            & PEZ & \acc{54.54}{1.84} & \acc{58.44}{2.35} & \acc{60.47}{2.68} & \acc{62.04}{1.61} & \acc{62.99}{2.26} \\
            & ProAPO & \acc{\underline{79.94}}{0.58} & \acc{\underline{80.17}}{0.33} & \acc{\underline{80.20}}{0.76} & \acc{\underline{80.36}}{0.52} & \acc{\underline{80.24}}{0.27} \\
            \rowcolor{blue!8}
            \cellcolor{white}
            \multirow{-6}{*}{\shortstack[l]{ViT-L: \\ DFN}}
            & \textbf{TPB (Ours)} & \acc{\mathbf{80.26}}{0.74} & \acc{\mathbf{80.88}}{0.71} & \acc{\mathbf{81.70}}{0.42} & \acc{\mathbf{82.24}}{0.53} & \acc{\mathbf{82.44}}{0.49} \\
            \midrule
            & ZS-CLIP & \multicolumn{5}{c}{74.96} \\
            \cmidrule(){2-7}
            & CoOp+EFT & \acc{71.42}{2.29} & \acc{73.54}{1.52} & \acc{77.02}{1.15} & \acc{79.53}{0.56} & \acc{\underline{81.45}}{0.35} \\
            & PromptSRC+EFT & \acc{74.89}{1.40} & \acc{76.04}{1.40} & \acc{\underline{79.10}}{0.79} & \acc{\mathbf{80.62}}{0.74} & \acc{\mathbf{82.08}}{0.38} \\
            \cmidrule(){2-7}
            & PEZ & \acc{51.83}{2.01} & \acc{55.67}{2.12} & \acc{58.35}{2.23} & \acc{61.90}{1.91} & \acc{63.26}{1.23} \\
            & ProAPO & \acc{\mathbf{78.33}}{0.68} & \acc{\underline{77.60}}{0.95} & \acc{78.62}{0.37} & \acc{78.45}{0.74} & \acc{78.77}{0.43} \\
            \rowcolor{blue!8}
            \cellcolor{white}
            \multirow{-6}{*}{\shortstack[l]{ViT-L: \\ EVA-02-CLIP}}
            & \textbf{TPB (Ours)} & \acc{\underline{77.94}}{0.75} & \acc{\mathbf{79.15}}{0.77} & \acc{\mathbf{79.36}}{0.70} & \acc{\underline{80.50}}{0.46} & \acc{80.42}{0.34} \\
            \midrule
            & ZS-CLIP & \multicolumn{5}{c}{76.82} \\
            \cmidrule(){2-7}
            & CoOp+EFT & \acc{74.11}{2.17} & \acc{76.13}{1.42} & \acc{78.88}{0.81} & \acc{80.70}{0.53} & \acc{\underline{82.17}}{0.29} \\
            & PromptSRC+EFT & \acc{76.55}{1.26} & \acc{77.40}{1.18} & \acc{80.03}{0.59} & \acc{\underline{81.18}}{0.82} & \acc{\mathbf{82.39}}{0.30} \\
            \cmidrule(){2-7}
            & PEZ & \acc{42.13}{2.61} & \acc{46.05}{2.53} & \acc{48.16}{3.06} & \acc{51.97}{1.60} & \acc{51.95}{2.30} \\
            & ProAPO & \acc{\mathbf{79.88}}{0.49} & \acc{\underline{80.02}}{0.47} & \acc{\underline{80.40}}{0.47} & \acc{80.33}{0.28} & \acc{80.46}{0.43} \\
            \rowcolor{blue!8}
            \cellcolor{white}
            \multirow{-6}{*}{\shortstack[l]{ViT-L: \\ CLIPA-v2}}
            & \textbf{TPB (Ours)} & \acc{\underline{79.41}}{0.19} & \acc{\mathbf{80.14}}{0.53} & \acc{\mathbf{80.78}}{0.54} & \acc{\mathbf{81.57}}{0.28} & \acc{81.75}{0.30} \\
            \midrule
            \rowcolor{gray!15}
            \multicolumn{7}{c}{\textit{Target Models: ViT-H family}}\\
            \midrule
            & ZS-CLIP & \multicolumn{5}{c}{76.51} \\
            \cmidrule(){2-7}
            & CoOp+EFT & \acc{76.18}{1.39} & \acc{77.73}{0.75} & \acc{79.84}{0.51} & \acc{\underline{81.48}}{0.42} & \acc{\mathbf{82.88}}{0.17} \\
            & PromptSRC+EFT & \acc{77.80}{0.78} & \acc{78.88}{0.74} & \acc{\underline{80.56}}{0.35} & \acc{81.45}{0.42} & \acc{82.53}{0.28} \\
            \cmidrule(){2-7}
            & PEZ & \acc{48.55}{1.94} & \acc{53.71}{2.12} & \acc{56.95}{1.60} & \acc{59.12}{1.50} & \acc{60.55}{2.05} \\
            & ProAPO & \acc{\mathbf{79.92}}{0.42} & \acc{\underline{80.11}}{0.63} & \acc{80.41}{0.44} & \acc{80.44}{0.50} & \acc{80.72}{0.54} \\
            \rowcolor{blue!8}
            \cellcolor{white}
            \multirow{-6}{*}{\shortstack[l]{ViT-H: \\ OpenCLIP}}
            & \textbf{TPB (Ours)} & \acc{\underline{79.86}}{0.78} & \acc{\mathbf{80.21}}{0.54} & \acc{\mathbf{81.63}}{0.35} & \acc{\mathbf{82.26}}{0.42} & \acc{\underline{82.82}}{0.59} \\
            \midrule
            & ZS-CLIP & \multicolumn{5}{c}{81.51} \\
            \cmidrule(){2-7}
            & CoOp+EFT & \acc{25.11}{5.02} & \acc{27.94}{4.28} & \acc{34.17}{3.85} & \acc{37.54}{2.28} & \acc{42.09}{1.37} \\
            & PromptSRC+EFT & \acc{29.12}{3.81} & \acc{32.23}{3.61} & \acc{39.36}{4.66} & \acc{43.38}{2.41} & \acc{47.54}{1.71} \\
            \cmidrule(){2-7}
            & PEZ & \acc{56.60}{1.46} & \acc{60.99}{2.26} & \acc{64.50}{2.39} & \acc{66.28}{1.70} & \acc{66.76}{1.78} \\
            & ProAPO & \acc{\mathbf{84.41}}{0.70} & \acc{\underline{84.01}}{0.63} & \acc{\underline{84.32}}{0.43} & \acc{\underline{84.32}}{0.44} & \acc{\underline{84.62}}{0.44} \\
            \rowcolor{blue!8}
            \cellcolor{white}
            \multirow{-6}{*}{\shortstack[l]{ViT-H: \\ DFN}}
            & \textbf{TPB (Ours)} & \acc{\underline{83.86}}{0.50} & \acc{\mathbf{84.61}}{0.45} & \acc{\mathbf{85.30}}{0.34} & \acc{\mathbf{86.02}}{0.32} & \acc{\mathbf{85.83}}{0.49} \\
            \midrule
            & ZS-CLIP & \multicolumn{5}{c}{78.50} \\
            \cmidrule(){2-7}
            & CoOp+EFT & \acc{76.85}{1.80} & \acc{78.17}{1.04} & \acc{80.59}{0.69} & \acc{82.21}{0.51} & \acc{\underline{83.57}}{0.32} \\
            & PromptSRC+EFT & \acc{78.65}{1.06} & \acc{79.32}{0.99} & \acc{81.52}{0.54} & \acc{\underline{82.60}}{0.63} & \acc{83.52}{0.34} \\
            \cmidrule(){2-7}
            & PEZ & \acc{43.07}{2.68} & \acc{47.17}{2.51} & \acc{49.63}{3.10} & \acc{52.30}{1.81} & \acc{53.13}{1.82} \\
            & ProAPO & \acc{\mathbf{81.80}}{0.64} & \acc{\mathbf{81.71}}{0.52} & \acc{\underline{82.10}}{0.73} & \acc{82.16}{0.65} & \acc{82.27}{0.44} \\
            \rowcolor{blue!8}
            \cellcolor{white}
            \multirow{-6}{*}{\shortstack[l]{ViT-H: \\ CLIPA-v2}}
            & \textbf{TPB (Ours)} & \acc{\underline{81.50}}{0.48} & \acc{\underline{81.68}}{0.95} & \acc{\mathbf{82.94}}{0.43} & \acc{\mathbf{83.43}}{0.39} & \acc{\mathbf{83.83}}{0.49} \\
            \midrule
            & ZS-CLIP & \multicolumn{5}{c}{78.47} \\
            \cmidrule(){2-7}
            & CoOp+EFT & \acc{72.61}{2.61} & \acc{74.69}{1.96} & \acc{78.07}{1.04} & \acc{80.37}{0.53} & \acc{82.39}{0.21} \\
            & PromptSRC+EFT & \acc{75.80}{2.03} & \acc{77.64}{1.87} & \acc{80.65}{0.63} & \acc{81.70}{0.96} & \acc{\underline{83.05}}{0.38} \\
            \cmidrule(){2-7}
            & PEZ & \acc{39.64}{1.86} & \acc{44.14}{2.46} & \acc{47.28}{2.23} & \acc{48.20}{1.69} & \acc{49.96}{1.70} \\
            & ProAPO & \acc{\mathbf{82.14}}{0.73} & \acc{\underline{81.78}}{0.74} & \acc{\underline{82.20}}{0.63} & \acc{\underline{82.10}}{0.64} & \acc{81.94}{0.34} \\
            \rowcolor{blue!8}
            \cellcolor{white}
            \multirow{-6}{*}{\shortstack[l]{ViT-H: \\ MetaCLIP2}}
            & \textbf{TPB (Ours)} & \acc{\underline{81.71}}{0.54} & \acc{\mathbf{81.96}}{0.68} & \acc{\mathbf{83.10}}{0.38} & \acc{\mathbf{83.66}}{0.59} & \acc{\mathbf{84.46}}{0.40} \\
            \bottomrule
        \end{tabular}
        }
    \end{table}
}

\subsection{Full target-wise table of boosting \vs. augmentation ablation.}
\Cref{tab:ma_full} provides the full target-model-wise results of the fixed-exposure ablation, where we keep the total exposure budget fixed as $M \cdot a = 200$ and vary the allocation between the number of boosting rounds $M$ and the augmentation factor $a$.
Under the same exposure budget, allocating more budget to boosting rounds with moderate augmentation ($M/a=50/4$) achieves the best performance on almost all target models, with only one exception where $M/a=10/20$ is slightly better.
This consistent trend across heterogeneous target VLMs supports the role of iterative reweighting and ensemble construction, rather than augmentation alone, in learning transferable prompt ensembles.
\begin{table}[t]
    \caption{\textbf{Full model-wise results under a fixed exposure budget.} Top-1 accuracy (\%) in the 16-shot setting, averaged over 10 datasets excluding ImageNet. We fix $M\!\cdot\!a=200$.}
    \label{tab:ma_full}

    \centering
    \footnotesize
    \setlength{\tabcolsep}{4pt}

    \scalebox{0.85}{
    \begin{tabular}{l|c|cccc}
        \toprule
        \multirow{2}{*}{\textbf{Model}} & \multirow{2}{*}{\textbf{ZS-CLIP}} & \multicolumn{4}{c}{$M/a$} \\
        & & $1/200$ & $5/40$ & $10/20$ & $50/4$ \\
        \midrule

        \rowcolor{gray!15}
        \multicolumn{6}{c}{\textit{Source Model: OpenAI CLIP, ViT-B/32}} \\

        \midrule
        OpenAI CLIP & 60.10 & \acc{\mathbf{74.42}}{0.17} & \acc{72.94}{0.10} & \acc{73.32}{0.26} & \acc{\underline{73.36}}{0.47} \\

        \midrule
        \rowcolor{gray!15}
        \multicolumn{6}{c}{\textit{Target Models: ViT-L family}} \\

        \midrule
        CLIPA-v2 & 76.61 & \acc{81.19}{0.46} & \acc{81.32}{0.28} & \acc{\underline{81.75}}{0.29} & \acc{\mathbf{81.89}}{0.31} \\
        DFN & 76.85 & \acc{81.89}{0.29} & \acc{82.01}{0.09} & \acc{\underline{82.32}}{0.31} & \acc{\mathbf{82.52}}{0.53} \\
        SigLIP2 & 81.44 & \acc{84.60}{0.24} & \acc{85.24}{0.21} & \acc{\underline{85.49}}{0.05} & \acc{\mathbf{85.73}}{0.38} \\
        EVA02 & 74.54 & \acc{\underline{80.78}}{0.35} & \acc{80.58}{0.63} & \acc{\mathbf{81.10}}{0.12} & \acc{80.38}{0.36} \\
        OpenCLIP & 73.23 & \acc{79.23}{0.30} & \acc{79.74}{0.16} & \acc{\underline{80.01}}{0.28} & \acc{\mathbf{80.68}}{0.37} \\
        \midrule
        \rowcolor{gray!15}
        \multicolumn{6}{c}{\textit{Target Models: ViT-H family}} \\

        \midrule
        CLIPA-v2 & 78.27 & \acc{82.54}{0.41} & \acc{83.03}{0.31} & \acc{\underline{83.33}}{0.06} & \acc{\mathbf{84.00}}{0.52} \\
        DFN & 81.37 & \acc{85.62}{0.50} & \acc{85.66}{0.56} & \acc{\underline{85.78}}{0.26} & \acc{\mathbf{86.04}}{0.52} \\
        MetaCLIP2 & 78.27 & \acc{82.77}{0.28} & \acc{83.73}{0.11} & \acc{\underline{84.09}}{0.57} & \acc{\mathbf{84.70}}{0.42} \\
        OpenCLIP & 76.47 & \acc{81.53}{0.19} & \acc{82.46}{0.15} & \acc{\underline{82.71}}{0.29} & \acc{\mathbf{83.20}}{0.63} \\
        
        \bottomrule
    \end{tabular}
    }
\end{table}

\end{document}